\RequirePackage{iftex}
\documentclass[nolineno]{ieeeaccess}
\ifPDFTeX
\usepackage[T1]{fontenc}
\usepackage[utf8]{inputenc}
\usepackage{mathptmx,helvet}
\else
\usepackage{fontspec}
\fi
\usepackage{amsmath,amssymb}
\usepackage{graphicx,booktabs,tabularx,array,multirow}
\usepackage{cite,url,etoolbox,fancyhdr}
\usepackage[hidelinks]{hyperref}
\usepackage{placeins}
\usepackage{balance}
\newcolumntype{P}[1]{>{\raggedright\arraybackslash}p{#1}}
\newcolumntype{Y}{>{\raggedright\arraybackslash}X}

\renewcommand{\arraystretch}{1.12}
\graphicspath{{figures/}}
\makeatletter
\AtBeginDocument{\typeout{LAYOUTHEIGHT \the\textheight}}
\apptocmd{\@endfloatbox}{\typeout{FLOATSIZE \@captype\space \number\csname c@\@captype\endcsname\space HEIGHT \the\ht\@currbox\space DEPTH \the\dp\@currbox}}{}{}

\long\def\@makecaption#1#2{\vskip6pt{\sffamily\fontsize{8}{10}\selectfont\raggedright\textbf{#1.}\enspace #2\par}\vskip5pt}
\patchcmd{\@maketitle}{\vss\doifont Digital Object Identifier\space\@doi\vss}{}{}{}
\def\ps@titlepage{\def\@oddhead{}\def\@evenhead{}\def\@oddfoot{\hfil\footerpagefont\thepage}\let\@evenfoot\@oddfoot}
\def\ps@headings{\def\@oddhead{\headerfont Recursive Self-Improvement in Physical AI\hfil}\let\@evenhead\@oddhead\def\@oddfoot{\hfil\footerpagefont\thepage}\let\@evenfoot\@oddfoot}
\def\EOD{\global\eoddefinedtrue}
\makeatother
\title{Recursive Self-Improvement in Physical AI: Mechanisms, Evidence, and Evaluation}
\newcommand{\authorblockpar}{\par}
\renewcommand{\titlefont}{\sffamily\fontencoding{T1}\fontseries{b}\fontsize{20}{23}\selectfont}
\renewcommand{\authorfont}{\normalfont\rmfamily\fontsize{11}{14}\selectfont}
\renewcommand{\addressfont}{\normalfont\rmfamily\itshape\fontsize{9}{11}\selectfont\centering}
\author{\centering
Xiaofeng~Han\textsuperscript{b,a,*},
Kaize~Luo\textsuperscript{c,*},
Zenghuang~Fu\textsuperscript{b,a,*},
Zichong~Zhu\textsuperscript{f},
Zhe~Feng\textsuperscript{b,a},\\[2pt]
Changwei~Wang\textsuperscript{d,e},
Rongtao~Xu\textsuperscript{a,b,\textdagger},
Weiliang~Meng\textsuperscript{a,b},
Xiaopeng~Zhang\textsuperscript{a,b}\authorblockpar\vspace{2mm}}
\address[a]{The State Key Laboratory of Multimodal Artificial Intelligence Systems, Institute of Automation, Chinese Academy of Sciences, China\authorblockpar}
\address[b]{School of Artificial Intelligence, University of Chinese Academy of Sciences, China\authorblockpar}
\address[c]{Southern University of Science and Technology, China\authorblockpar}
\address[d]{Key Laboratory of Computing Power Network and Information Security, Ministry of Education; Shandong Computer Science Center, Qilu University of Technology (Shandong Academy of Sciences), China\authorblockpar}
\address[e]{Shandong Provincial Key Laboratory of Computing Power Internet and Service Computing, Shandong Fundamental Research Center for Computer Science, China\authorblockpar}
\address[f]{Shenzhen Institutes of Advanced Technology, Chinese Academy of Sciences, China\authorblockpar}
\corresp{}
\history{}
\doi{}
\tfootnote{}
\hypersetup{pdfauthor={Xiaofeng Han, Kaize Luo, Zenghuang Fu, Zichong Zhu, Zhe Feng, Changwei Wang, Rongtao Xu, Weiliang Meng, Xiaopeng Zhang},pdftitle={Recursive Self-Improvement in Physical AI: Mechanisms, Evidence, and Evaluation},pdfsubject={A structured narrative survey},pdfkeywords={recursive self-improvement, physical AI, robot learning, embodied AI, evaluation}}

\makeatletter
\AtEndDocument{\@for\layoutlabel:=sec:intro,fig:concept,tab:related-surveys,sec:foundations,fig:boundary,tab:evidence-levels,fig:levels,fig:lineages,sec:mechanisms,fig:carriers,fig:worldroles,sec:recursion,fig:targets,sec:physical,fig:continuity,fig:failure,sec:evaluation,tab:evidence-fields,tab:assessments,fig:benchmarks,tab:robomemarena,tab:robodojo-performance,tab:robodojo-evolution,tab:reported-gains,tab:resource-protocols,fig:conditions,fig:swap,tab:improver-controls,sec:directions,sec:conclusion\do{\typeout{LAYOUTCHECK \layoutlabel\space PAGE \getpagerefnumber{\layoutlabel}}}}
\makeatother

\begin{document}
\begin{abstract}
Embodied systems can update policies, world models, skills, and executable programs from deployment experience, allowing one interaction to benefit later tasks. When a system also revises and reuses its data-acquisition, feedback, or learning procedures, experience changes the process that produces capability, introducing recursive self-improvement. This paper surveys research in physical AI through the connection between persistent capability and its improvement process. We first explain the functional distinction between learner updates and improver revisions, then compare policy learning, imagined experience, memory and skill accumulation, and executable harnesses. These approaches connect to revisions of curricula, rewards, and research programs in different ways. Contact validity, recovery, retention, and resource costs determine whether the resulting gains accumulate during continued deployment. An L0--L3 rubric distinguishes task-local adaptation, persistent capability gain, inherited process revision, and verified improver gain. Representative benchmarks and within-study contrasts relate these distinctions to existing results. We further propose common-data and interactive comparisons that evaluate reference and revised improvers from shared learner starts, separating the use of experience from its acquisition. This perspective connects the diverse methods, physical operating conditions, and evaluation evidence underlying embodied RSI. A comprehensive list of studies in this survey is available at \href{https://github.com/nkd-lkz/physical-ai-rsi-survey}{https://github.com/nkd-lkz/physical-ai-rsi-survey}.
\end{abstract}
\begin{keywords}
physical AI; recursive self-improvement; robot learning; world models; skill memory; agent harness; evaluation.
\end{keywords}
\maketitle
\begingroup
\renewcommand{\thefootnote}{}
\footnotetext{\normalfont\rmfamily\fontsize{9}{11}\selectfont\textsuperscript{*}Equal contribution.\\
\textsuperscript{\textdagger}Corresponding authors.}
\endgroup
\pagestyle{headings}
\section{Introduction}
\label{sec:intro}

Robots encounter new objects, contact conditions, and task combinations after deployment. A failed attempt exposes a capability gap and can also supply experience for the next improvement. Failed grasps can inform policy updates, human corrections can become reusable skills, and recurring failures can guide further practice. Self-improvement methods incorporate this experience into continued learning so that later tasks benefit from earlier interaction. RoboCat and RECAP use robot experience to update policies, while Learning While Deploying extends the reuse of deployment experience across robot fleets. \cite{W005,W003,W004}

As foundation models generate programs, interpret failures, and organize experiments, the editable parts of a robot system extend beyond its action policy to the learning process. ASPIRE and SHAPER retain changes to skills and harnesses, while ENPIRE exposes policy-development programs to agent-generated revision. This enables a further form of self-improvement: experience changes data selection, feedback construction, or training code, and the revised process produces the next update. We use mechanistic recursive self-improvement (RSI) for such system-generated process revisions that are retained and used in subsequent learning. Experience thereby changes how the system acquires new capabilities. \cite{W027,W029,W028}

Current work approaches self-improvement through policy training, world models, skill libraries, harnesses, and automated experimentation. The same terminology often refers to changes at different levels. Continued practice can improve a policy, revised reward code can change its learning objective, and a more effective experimental procedure can improve subsequent learning efficiency. Connecting these approaches requires distinguishing changes in capability, changes in the learning process, and the additional gains produced by those process changes. Embodied systems must also obtain valid feedback, recover from failures, and retain existing capabilities within limited robot time and human assistance.

\begin{figure*}[t]
\centering\includegraphics[width=\textwidth]{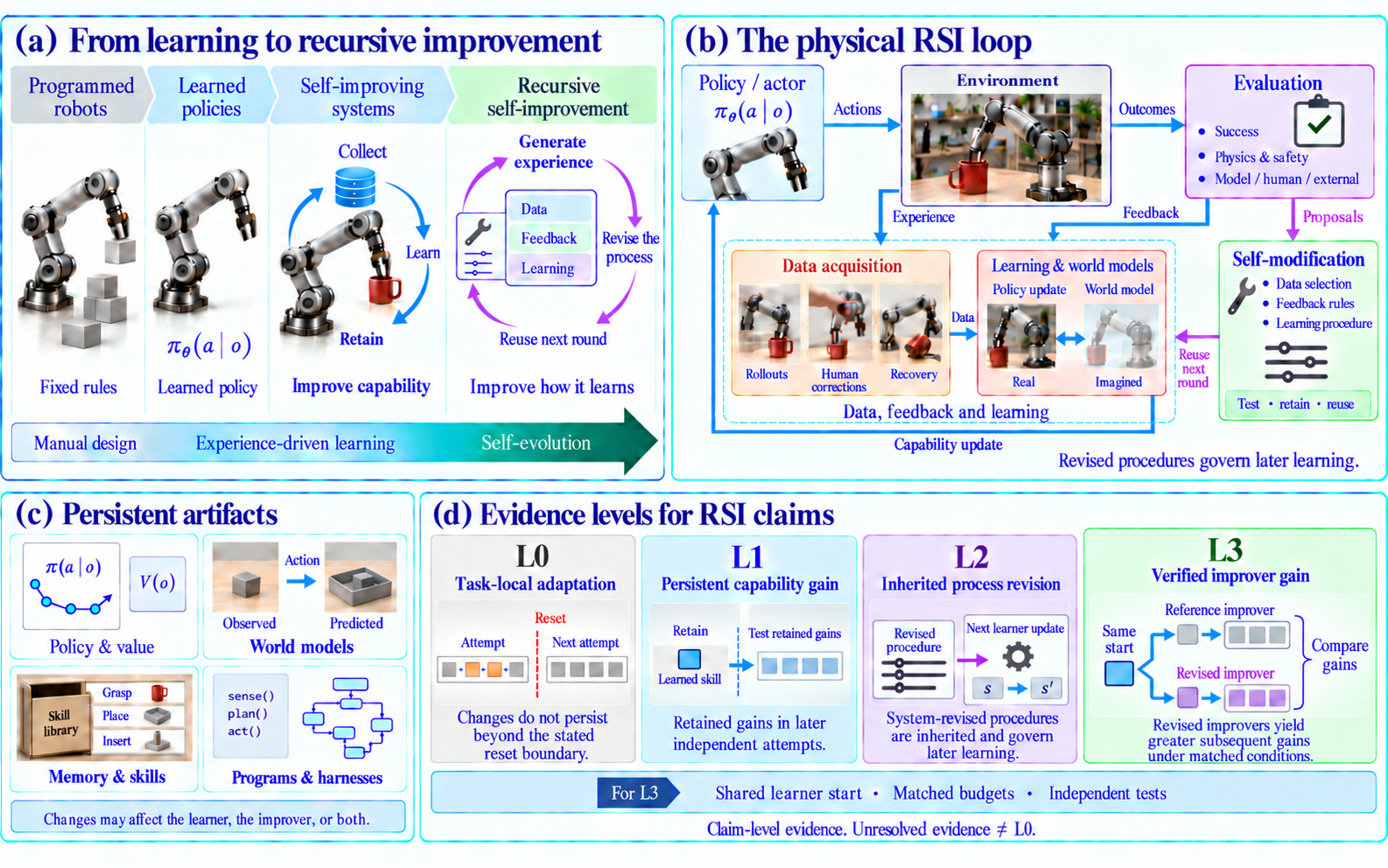}
\caption{Overview of the review framework for recursive self-improvement in physical AI. (a) Conceptual distinctions among programmed robots, learned policies, persistent self-improvement, and recursive improvement. (b) Physical experience and feedback support capability updates, while system-generated process revisions are retained and reused to govern later learning. (c) Persistent artifacts can modify the learner, the improver, or both. (d) The evidence rubric distinguishes task-local adaptation (L0), persistent capability gain (L1), inherited process revision (L2), and verified improver gain (L3). L3 requires greater subsequent gains from shared learner starts under matched resource budgets and independent tests. The levels specify evidence requirements for individual claims. Comparisons are schematic, and robot scenes are illustrative renderings.}
\label{fig:concept}
\end{figure*}

With robot learning and deployment as the main setting, this paper connects two strands of self-improvement in physical AI: how experience becomes persistent capability, and how it changes the process that produces capability. The first covers policies and values, world models, memory and skills, and executable harnesses. The second concerns revisions to experience acquisition, feedback construction, and learning procedures. Figure~\ref{fig:concept} summarizes their relationship. The L0--L3 evidence rubric connects advances in these mechanisms to the gains established by their evaluations.

This paper makes four contributions:
\begin{itemize}
\item \textbf{Conceptual distinctions and evidence criteria.} We distinguish task-local adaptation, persistent capability gain, inherited process revision, and verified improver gain. The L0--L3 rubric connects these claims to observable evidence for a specific system configuration and protocol.
\item \textbf{A synthesis of persistent and recursive mechanisms.} We organize methods by what experience changes, what is retained, and how the retained change influences subsequent learning. This functional analysis links updates to the learner with revisions to data acquisition, feedback, learning procedures, and update control.
\item \textbf{Physical conditions and controlled evaluation.} We connect contact validity, recovery, retention, and resource use to the interpretation of improvement. Common-data and interactive comparisons distinguish the use of experience from its acquisition when assessing reference and revised improvers.
\item \textbf{Evidence-based research priorities.} We derive directions for grounded feedback, lasting gains, reliable recovery, and efficient improvement from the gaps between implemented mechanisms and demonstrated benefits.
\end{itemize}

Related surveys provide complementary perspectives on self-evolving agents, embodied learning, and world--action models.\cite{W093,W100,W105} Embodied self-evolution has been organized around memory, tasks, world models, embodiment, and model design, including changes to optimization and evaluation procedures.\cite{W095} Code-based robot-learning surveys distinguish execution feedback, persistent memory, and program search,\cite{W096} while continual-learning research develops retention and transfer metrics alongside practical robotics constraints.\cite{lesort2020continual} Recent humanoid surveys connect post-deployment mechanisms and meta-updates to safety constraints and lifetime trajectory evaluation.\cite{REC_HumanoidSelfEvo2026} General RSI surveys also cover embodied applications and controlled comparisons of original and revised improvement mechanisms under matched budgets.\cite{W092}

These surveys establish the components of self-evolution. This paper connects the technical approaches through the relationship between experience, retained artifacts, and subsequent learning: how policies, models, skills, and code acquire persistent value, and when those same components participate in revising the improvement process. Decision-centered world-model evaluation similarly relates component quality to its effects on planning and policy optimization. Table~\ref{tab:related-surveys} summarizes the related perspectives. This mechanism analysis provides the basis for the evidence rubric and physical evaluation conditions developed here. \cite{REC_WorldEvalPosition2026}

\begin{table*}[tp]
\caption{Relation to closely related surveys. Comparisons concern organizing perspectives and overlapping coverage.}
\label{tab:related-surveys}
\centering\small
\setlength{\tabcolsep}{4pt}
\begin{tabularx}{\textwidth}{P{0.25\textwidth}YY}
\toprule
\textbf{Survey} & \textbf{Organizing focus} & \textbf{Connection to this paper}\\\midrule
\textit{Self-evolving Embodied AI}\cite{W095} & Memory, tasks, world models, embodiment, and model evolution & Broad embodied framework, including changes to optimization and evaluation procedures.\\[4pt]
\textit{Weights or Skills?}\cite{W096} & Execution feedback, persistent memory, and code-based skill search & Operational distinctions between task-local repair and reusable capability.\\[4pt]
\textit{Self-Evolving AI for Humanoids}\cite{REC_HumanoidSelfEvo2026} & Post-deployment mechanisms, meta-updates, safety, and lifetime trajectories & Physical evolution and trajectory evaluation, complementing separate tests of process inheritance and improver gain.\\[4pt]
\textit{The Last AI Built by Humans}\cite{W092} & RSI autonomy levels, application domains, and controlled improver evaluation & Direct RSI perspective; autonomy and evidential strength characterize different aspects of improvement.\\[4pt]
\textit{Continual Learning for Robotics}\cite{lesort2020continual} & Continual-learning strategies, retention, transfer, and robotics constraints & Foundations for persistent-capability assessment and physical evaluation conditions.\\
\bottomrule
\end{tabularx}
\end{table*}

The remainder of the review follows this progression. Section II establishes the conceptual framework and evidence criteria. Sections III and IV examine persistent capability updates and inherited revisions to the improvement process, respectively. Section V analyzes the physical conditions for sustained improvement, and Section VI synthesizes the evidence and evaluation protocols. Section VII develops research directions from the resulting gaps, and Section VIII concludes.

\section{Foundations and Conceptual Framework}
\label{sec:foundations}

Better performance after interaction can arise from adaptation within the current episode, capability retained across tasks, or changes to the learning process itself. We distinguish these cases through the object of change, its retention boundary, and its later use. The functional distinction between learner and improver explains the mechanism; capability gain, mechanistic recursion, and improver gain identify the empirical claims; and the L0--L3 rubric connects those claims to observable evidence requirements. These distinctions also help interpret the convergence of several research traditions, providing a common basis for comparing their approaches.

\subsection{Learner state and improvement process}

A robot learning to grasp provides an intuitive distinction between capability and the improvement process. Updating its grasping policy from failed trajectories changes its retained action capability. If the system also identifies missing slip experience, revises its sampling or feedback procedure, and reuses that revision in later training, it changes the process that produces capability. Policies, predictive models, and reusable skills constitute the learner state $x$. Procedures that select experience, construct feedback, produce updates, and accept changes constitute the improver $\varphi$. We use self-evolution for experience-driven, system-generated changes retained for later use; recursive self-improvement additionally involves revision of the improver. These updates can be written as

\begin{equation}
\begin{aligned}
x_{t+1}&=U_{\varphi_t}(x_t,D_t),\\
\varphi_{t+1}&=R(\varphi_t,x_t,D_t).
\end{aligned}
\label{eq:learner_improver}
\end{equation}
Here, $D_t$ is the interaction experience available at round $t$. The first equation describes how improver $\varphi_t$ produces a new learner state through update operation $U$. The second describes how revision operation $R$ produces a version of the improver for later use. Fixed-process learning maintains $\varphi_{t+1}=\varphi_t$. The recursive connection occurs when the version produced by the second equation returns to the first and governs later capability updates. Its benefit is evaluated through the resulting learning performance.

The learner--improver boundary follows the use of a component. A world model can plan actions for the robot or generate experience for its training procedure. The former supports task execution directly; the latter participates in capability acquisition. An executable skill can likewise encode a grasp or an experimental schedule. Weights, memory, and code therefore do not correspond to fixed evidence levels. The use of an edited component determines whether it changes how the robot acts, how it continues to learn, or both.

This functional distinction also determines the experimental comparisons developed later. When two improvers are compared, the learner starts with the same policy, memory, and normalization statistics. Update code, improver-specific parameters, and meta-memory move with the improver. If a world model is shared, its interface must identify the function being replaced. Subsequent differences can then be related to the improvement process rather than unequal initial capabilities. An improvement round includes an opportunity to acquire experience and propose updates; its duration depends on the protocol.

Recursion can occur within a component of the improver. For example, a system may generate and retain revised reward code, then use it to train the next policy. This establishes inheritance from reward design to subsequent learning even when the outer code generator $R$ remains fixed. Adjusting control outputs, advancing subgoals, or accumulating optimizer statistics under an unchanged procedure instead executes that procedure. The distinction is whether later learning uses a process component that the system has revised and retained.

Meta-learning provides a useful boundary case for recursive improvement because the learned object can occupy different functional roles. MAML learns the initialization from which a fixed adaptation rule operates. In RL², outer-loop training encodes a fast learning algorithm in recurrent-network weights, while hidden-state changes implement adaptation during a trial. Learned optimizers and discovered reinforcement-learning algorithms also parameterize update rules, and meta-gradient reinforcement learning adjusts learning-objective parameters using subsequent experience. The distinction therefore lies in what is carried forward and how it governs later updates. A learned initialization, an adaptive hidden state, and an inherited update rule should receive separate evidence codes even when all improve adaptation. Testing an improver additionally requires comparing subsequent learning from a controlled learner start.\cite{F_MAML,F_RL2,F_L2L,F_METAGRAD,F_LPG}

\subsection{Three empirical claims}

These updates correspond to three empirical objects. Capability gain concerns whether the robot performs better after an update. Mechanistic recursion concerns whether the system revises and continues using the process that produces updates. Improver gain concerns whether the revised process produces better subsequent learning. In the grasping example, higher success supports a capability gain, an inherited sampling revision establishes a process change, and better later policies trained from matched starts and resources support an advantage of that revision. A process revision can be effective or ineffective, so its mechanism and benefit require separate observations.

Persistence must be measured across a declared boundary. State retained between attempts within one episode differs from state retained after an independent reset, across tasks, or across deployments. Zeva and in-context world modeling are useful boundary cases because their adaptation relies on interaction history without establishing the same form of long-term accumulation as a growing skill repository. Test-time parameter updating likewise requires an explicit account of what is reset before the next task. \cite{W048,W049,W050,W068,W099}

\begin{figure*}[t]
\centering\includegraphics[width=0.90\textwidth]{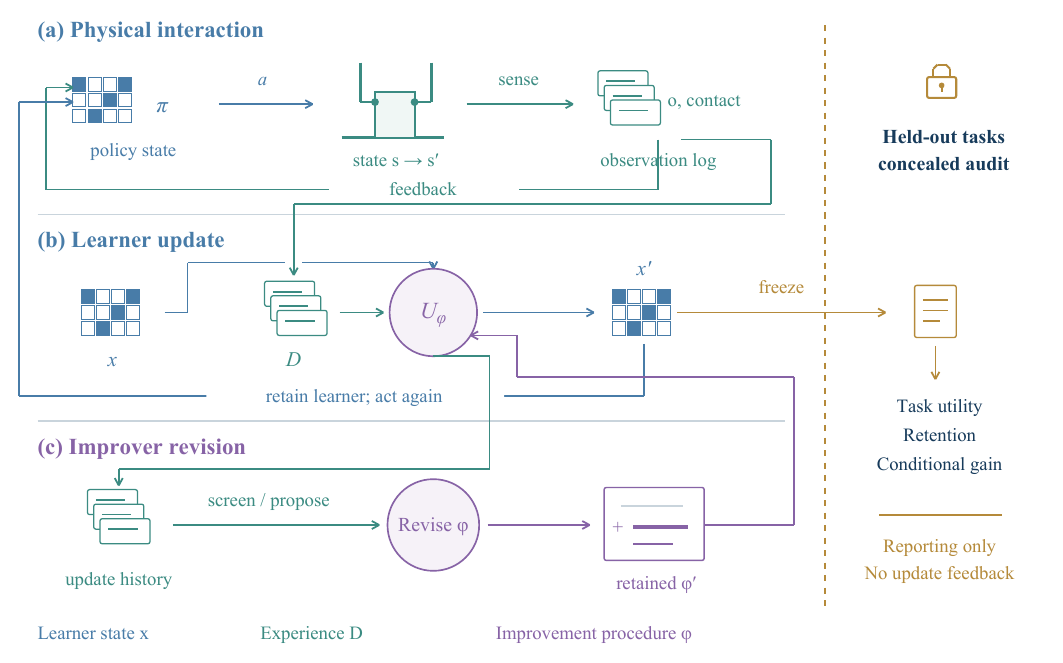}
\caption{Information boundaries between physical interaction, learner updates, and improver revision. Observations supply experience for learner updates; system-generated procedure changes are retained and used in later update rounds. Development screening can inform these revisions. A concealed reporting audit evaluates frozen candidates on independent held-out outcomes and supplies no feedback to the updates being assessed. Grids, contact geometry, records, and program differences are schematic representations.}
\label{fig:boundary}
\end{figure*}

Experience thus supports two connected operations: training the learner and informing revisions to its improver. Figure~\ref{fig:boundary} shows these paths and their connection to physical interaction. Independent outcome tests remain outside the update loop so that their measurements provide an external assessment of improvement.

Current-attempt correction makes the reset boundary concrete. RoboMonkey samples and verifies candidate actions; CycleVLA predicts failure and backtracks at subtask boundaries; closed-loop multi-robot agents coordinate execution through outcome verification and semantic correction.\cite{REC_RoboMonkey2025,REC_CycleVLA2026,REC_MultiRobotLoop2026} These operations can improve an ongoing attempt. Their persistent training components, when present, should be recorded separately from the inference-time correction loop. The relevant observation is what survives when context, environment, and task initialization are reset for a later independent attempt.

\subsection{A progressive L0--L3 rubric}

\begin{figure*}[t]
\centering\includegraphics[width=\textwidth]{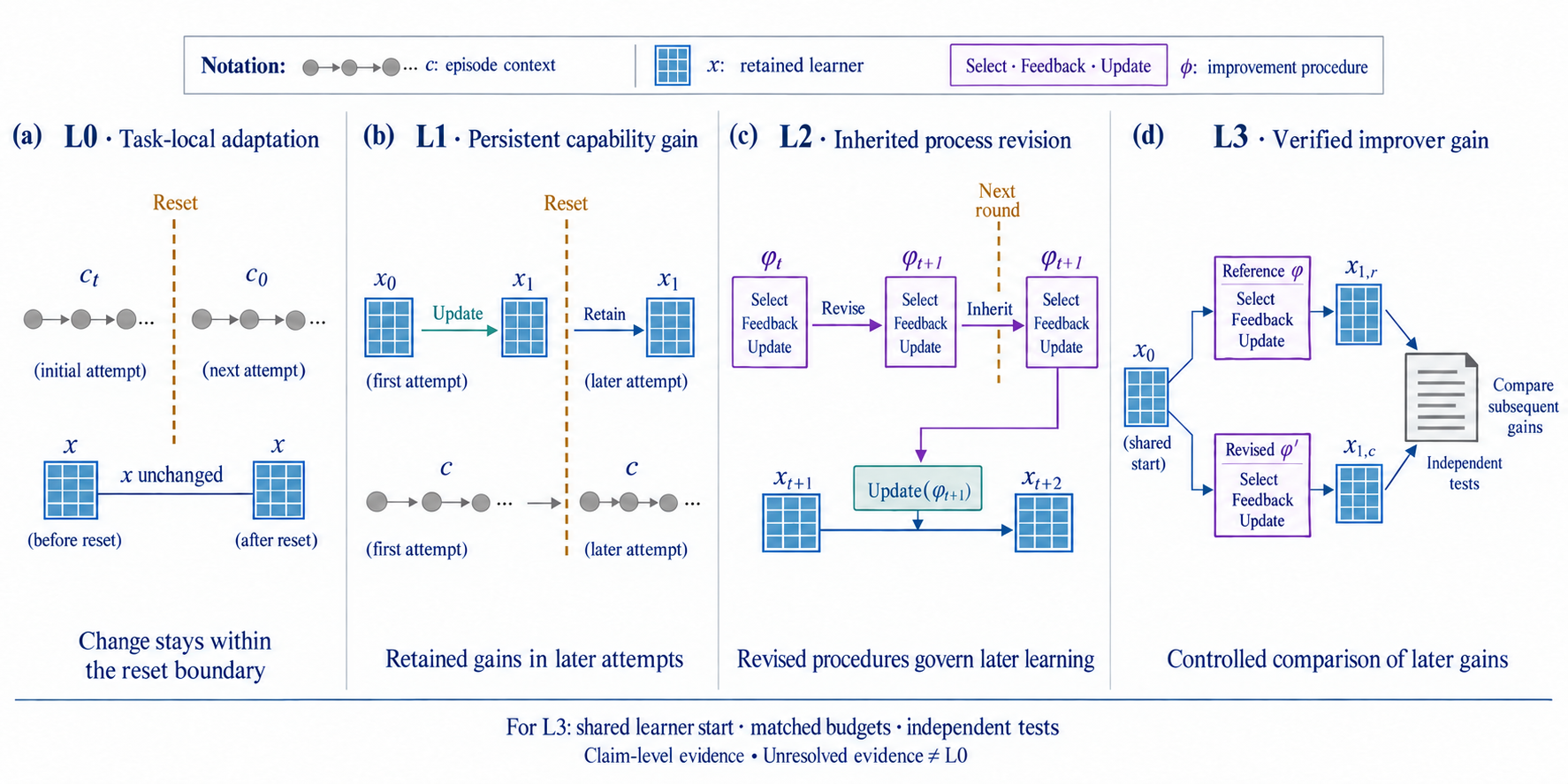}
\caption{Evidence requirements across a declared reset boundary. Episode context c, retained learner state x, and improvement procedure $\varphi$ have different persistence roles. L0 describes task-local changes that reset; L1 concerns retained capability gains in later independent attempts. L2 requires a system-generated process revision to be inherited and used in a later learner update. L3 requires greater subsequent gains under shared learner starts, matched resource budgets, and independent tests. The diagrams illustrate the criteria and corresponding evaluation conditions.}
\label{fig:levels}
\end{figure*}

We use four evidence levels, L0--L3, to characterize a specified system configuration and evaluation protocol. L0 concerns task-local adaptation within a reset boundary. L1 concerns retained capability gains in later independent attempts. L2 concerns system-generated process revisions that are inherited and govern subsequent updates. L3 concerns the revised improver's advantage in subsequent learning under controlled conditions. The levels progress from task-local change to persistent benefit, process inheritance, and improver gain. Figure~\ref{fig:levels} illustrates these distinctions; Table~\ref{tab:evidence-levels} states the operational criteria.

\begin{table}[tb]
\caption{Progressive claims and minimum evidence}
\label{tab:evidence-levels}

\centering\fontsize{8.5}{10.5}\selectfont
\setlength{\tabcolsep}{3pt}
\begin{tabular}{>{\raggedright\arraybackslash}p{0.09\columnwidth}>{\raggedright\arraybackslash}p{0.30\columnwidth}>{\raggedright\arraybackslash}p{0.53\columnwidth}}
\toprule
\textbf{Level} & \textbf{Operational claim} & \textbf{Minimum distinguishing evidence}\\\midrule
L0 & Task-local adaptation & Feedback changes the current attempt or episode; the adapted state is discarded at the declared reset.\\[3pt]
L1 & Persistent capability gain & Experience produces retained capability and benefits later independent attempts or tasks.\\[3pt]
L2 & Inherited process revision & System-generated revisions of an improvement component persist and govern subsequent learner updates.\\[3pt]
L3 & Verified improver gain & The revised improver yields greater subsequent gains from shared learner starts under matched resource limits and independent tests.\\[3pt]

\bottomrule
\end{tabular}
\end{table}

The progression measures accumulated evidence for a claim. Mechanism type, human involvement, physical setting, and evaluator independence remain separate descriptors. Level assignment follows the reported retained artifacts, reset boundaries, and evaluation controls. A reward model or dataset is analyzed through the system in which it is used. This separation supports meaningful comparisons between highly autonomous simulated systems and assisted physical systems without treating autonomy, embodiment, and recursion as a single score.

\subsection{Development through intersecting traditions}

The field develops through intersecting traditions rather than successive replacements. Model-based control and reinforcement learning made experience reusable through value estimates and predictive models; World Models and Dreamer exemplify this representational route. Open-ended learning, exemplified by POET, altered the distribution of problems encountered during learning. Program-generating agents later made skills and control logic explicit and reusable, as illustrated by Voyager and lifelong robot library learning. These traditions address different bottlenecks and continue to coexist. \cite{ha2018worldmodels,hafner2023dreamerv3,W076,W118,W040}

Recent work connects those traditions to foundation-model policies, learned feedback, and agent-managed experimentation. The important shift is an expansion from improving task solutions to making parts of the learning infrastructure editable. It does not establish a chronological sequence from L0 to L3: strong task-local adaptation can appear after persistent skill systems, and digital recursion can precede convincing physical validation. A historical account should therefore track mechanisms, feedback, and experimental scope separately. \cite{W003,W027,W028,W109,W111}

\begin{figure*}[p]
\centering
\includegraphics[width=\textwidth]{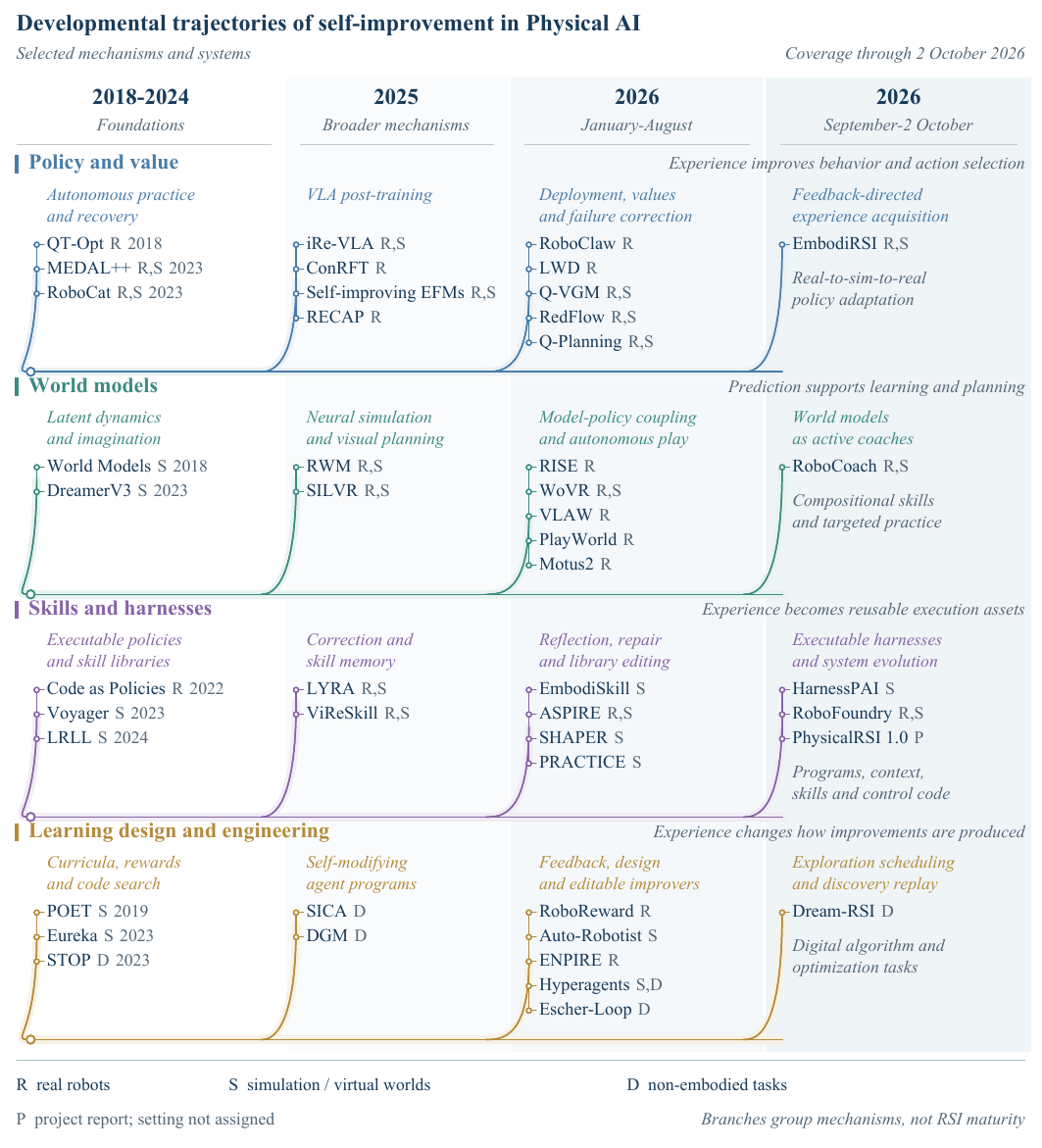}
\caption{Developmental trajectories of self-improvement in Physical AI. The 46 selected works are grouped by mechanism, with a separate column for recent work. Branches indicate thematic grouping, not causal inheritance or increasing RSI maturity. Intervals have unequal duration; within-period placement follows mechanism. R, S, and D denote reported real-robot, simulated/virtual, and non-embodied settings, not evidence strength. P denotes a project report without an assigned experimental-setting tag. Persistent updates and task gains alone do not establish improver gain.}
\label{fig:lineages}
\end{figure*}

A continuing theme in policy learning is expanding the experience available for improvement. Early autonomous practice and recovery enabled repeated sampling. Generalist robot policies reused experience across a wider range of tasks. VLA post-training combines value estimates, corrections, and learned feedback to make further use of successful and failed trajectories.\cite{liufu2026repo} These directions coexist with different data requirements and update interfaces. Figure~\ref{fig:lineages} places them within their related research traditions. \cite{kalashnikov2018qtopt,W070,W005,irevla2025,conrft2025,W009,W003,W037,W004,W025,W026,W010,N_EMBODIRSI}

The world-model lane distinguishes predictive simulation, imagined policy updates, and active coaching. The skills lane distinguishes reusable programs from memory management and execution-system evolution. Their convergence creates opportunities for shared artifacts, but it also makes the learner--improver boundary dependent on the artifact's actual use. \cite{ha2018worldmodels,hafner2023dreamerv3,li2025roboticworldmodel,W011,W001,W017,W006,W020,W002,N_ROBOCOACH,liang2022codeaspolicies,W118,W040,W033,W034,W041,W027,W029,W042,N_HARNESSPAI,N_ROBOFOUNDRY,N_PHYSICALRSI}

A second line of work treats tasks, rewards, morphology, and improvement programs as search objects. Curriculum and reward design change the conditions for policy learning, morphology design changes the fit between body and task, and editable agents bring the program that produces later solutions into the search. These approaches provide different entry points from improving a solution to improving how it is produced. Digital systems make program inheritance easier to examine; robotic systems additionally connect that inheritance to physical experience acquisition and execution. \cite{W076,W052,W109,robeyns2025sica,W111,W053,W082,W028,zhang2026hyperagents,W114,W119}

\section{Mechanisms of Persistent Self-Improvement}
\label{sec:mechanisms}

\begin{figure*}[t]
\centering\includegraphics[width=\textwidth]{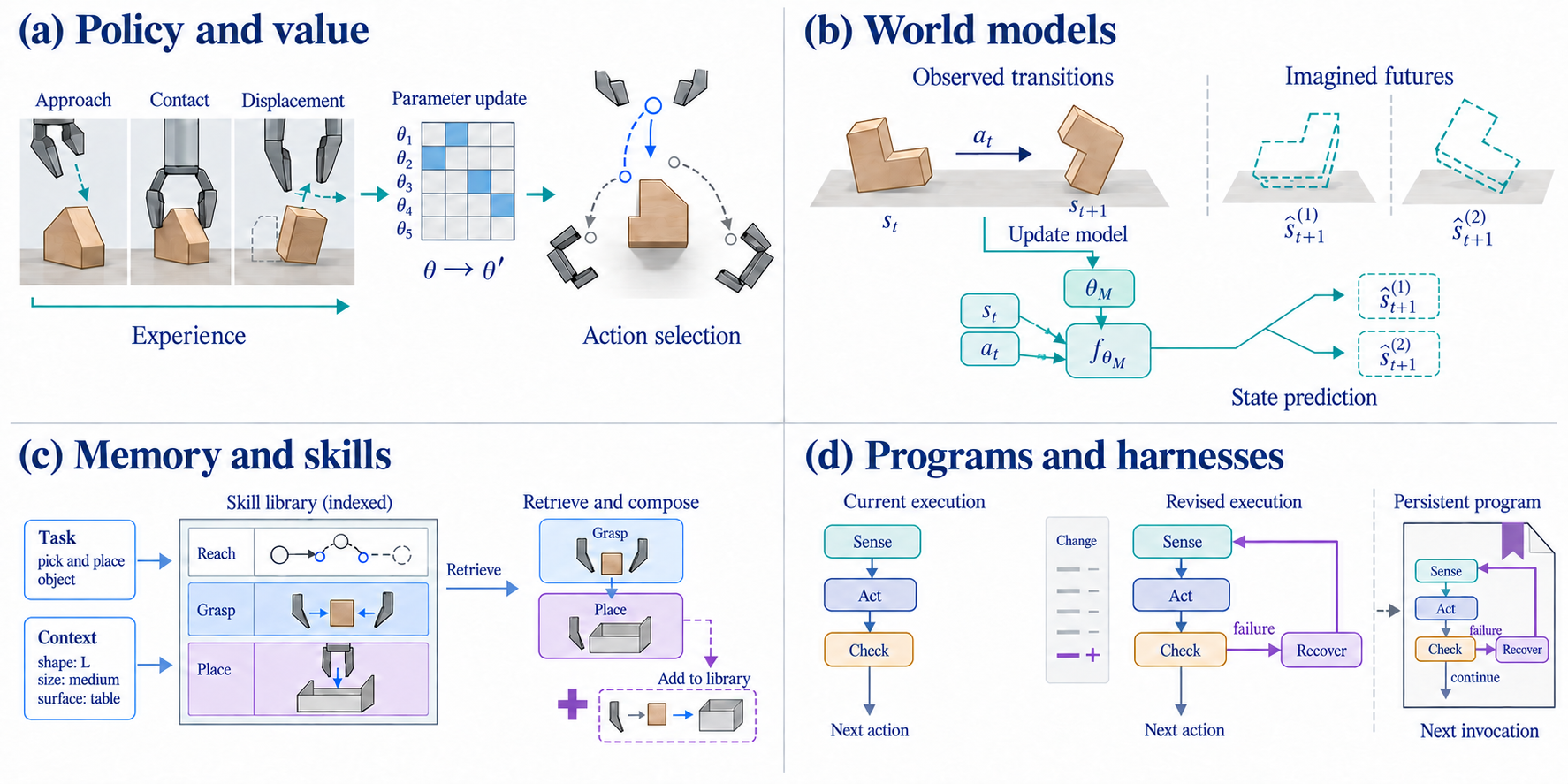}
\caption{Four carriers of persistent physical experience. Policy and value updates change parameters and action selection; world models retain predictive structure learned from observed transitions; indexed memory and skill libraries support retrieval and composition; programs and harnesses retain changes to execution logic, illustrated by a recovery branch. Each panel exposes the retained representation and its later use. A program change is classified by its functional role in execution or in producing later updates.}
\label{fig:carriers}
\end{figure*}

Persistent self-improvement turns experience from one interaction into capability that remains useful later. Different approaches retain that experience in different objects: policies and values influence action decisions, world models support prediction and planning, memory and skills provide retrievable knowledge, and executable programs and harnesses organize their use. Tracing experience through an update, its retention, and subsequent reuse (Figure~\ref{fig:carriers}) reveals the learning interfaces and scope of each approach. Several objects often change within the same system, making their division of labor and the effect of local updates on overall task capability central to the comparison.

\subsection{Policies and values}

Policy learning directly changes how a robot selects and executes actions.\cite{xu2025a0,zhang2026a1} For a vision--language--action model (VLA), post-deployment learning typically starts from pretrained capabilities and uses autonomous trajectories or corrective experience to adapt them to the task. RECAP uses value-derived advantage information to organize heterogeneous experience, while LWD aggregates deployment data from multiple robots. GR-RL and IG-RFT combine offline preparation with online interaction for difficult manipulation. These methods share the use of execution experience to update decision capability, with differences in experience organization, feedback, and update location. \cite{W003,W004,W012,W013}

The update location determines the interface through which experience changes actions. Q-Planning keeps the behavior policy frozen while learning to evaluate and select candidate actions; Q-VGM uses value signals to update an action expert. The former retains a new basis for selection, whereas the latter changes an action-generating component. Other methods focus on existing trajectories: CO-RFT improves their use in training, and RedFlow converts failures with matched references into action-level corrections. Persistent capability can therefore arise from newly collected data or from more effective learning with existing experience. \cite{W010,W025,W014,W026,lin2026evolvenav,zhang2025activevln}

Parameter-efficient updates further narrow the learning interface. DSRL trains the latent-noise input to a frozen diffusion policy, while RL Token trains compact actor--critic heads on a pretrained VLA representation.\cite{REC_DSRL2025,REC_RLToken2026} VLaRL learns latent-conditioned residual control in simulation before physical deployment; object-centric residual RL conditions corrections on object state and additionally studies reuse of improved rollouts.\cite{REC_VLaRL2026,REC_ObjectResidual2026} PLD uses residual learning to collect corrective trajectories and distills them into the generalist policy, whereas Hi-ORS combines online rejection sampling with human corrective trajectories.\cite{REC_PLD2025,REC_HiORS2025} Thus a frozen foundation model can coexist with a changing deployed learner. Residual-assisted execution, an independently learned controller, and a distilled successor policy should be evaluated as separate configurations, since each retains a different artifact and incurs a different runtime cost.

The intervention target determines both the scope of an update and the competence it can reach. RECAP changes the trained decision system using accumulated demonstrations, autonomous experience, and expert corrections; its task-specialization rounds refit value and policy models from pretrained checkpoints. Persistence therefore need not mean an uninterrupted chain of policy checkpoints: it can reside in accumulated experience and supervision. Q-Planning places adaptation elsewhere, learning a critic that weights action chunks proposed by a frozen behavior policy. Its smaller adaptation target can change executed behavior while leaving the underlying proposal mechanism fixed. A selector can improve the use of available proposals, but cannot establish competence outside their support merely by assigning different scores. Conversely, changing more policy parameters does not demonstrate better acquisition or feedback design. These methods should be compared through the complete deployed decision system, including action proposals, learned values, and data inherited between rounds. \cite{W003,W010}

Autonomous practice also depends on recognizing progress and finding useful starting behaviors. Self-Improving Embodied Foundation Models uses learned progress feedback to organize practice, whereas VLA jump-starting guides another learner with a foundation policy. These approaches use the foundation model as an evaluator or as a source of initial behavior, respectively, to support subsequent policy learning. \cite{W009,W016}

When a base policy struggles to obtain successful trajectories, auxiliary control can turn states near failure into learning opportunities. FIND combines autonomous execution with residual-policy learning, while SkillSpace Shooting obtains corrective experience through skills before retraining the policy. The benefit develops in two stages: assistance first improves collection, and the trained policy then absorbs the corrective experience. Removing the collection assistant during evaluation identifies which capabilities have transferred into the retained policy. \cite{N_FIND,N_SKILLSHOOT}

The post-training literature exposes several choices within policy-centered improvement. RIPT-VLA obtains advantages from sampled interactive rollouts and sparse outcomes; VLA-RL combines online optimization with learned process rewards; TGRPO combines trajectory-level and step-level credit with generated dense feedback.\cite{REC_RIPTVLA2025,REC_VLARL2025,REC_TGRPO2025} SimpleVLA-RL emphasizes the sampling and training infrastructure needed to scale this regime, whereas ReinFlow changes the stochastic representation of a flow policy so online reinforcement learning can optimize it.\cite{REC_SimpleVLARL2025,REC_ReinFlow2025} These approaches expose different bottlenecks: informative rollouts, credit assignment, feedback density, and the tractability of policy updates. Comparing final success alone merges these effects. A useful mechanism comparison also records the source of training feedback and which policy parameters are allowed to change.


\subsection{World models and imagined experience}

World models supplement physical trial and error by predicting the consequences of actions. A robot can compare predicted trajectories before execution or use generated experience to train its policy. The role of the model therefore determines where improvement occurs. RISE trains policies with imagined experience, SILVR improves visual planning from collected trajectories, and Motus2 connects policy, simulation, and value interfaces. Policy-training approaches convert predictions into retained action capability, planning approaches use predictions at decision time, and joint systems exchange information between these roles. \cite{W001,W002,W011}

\begin{figure*}[t]
\centering\includegraphics[width=\textwidth]{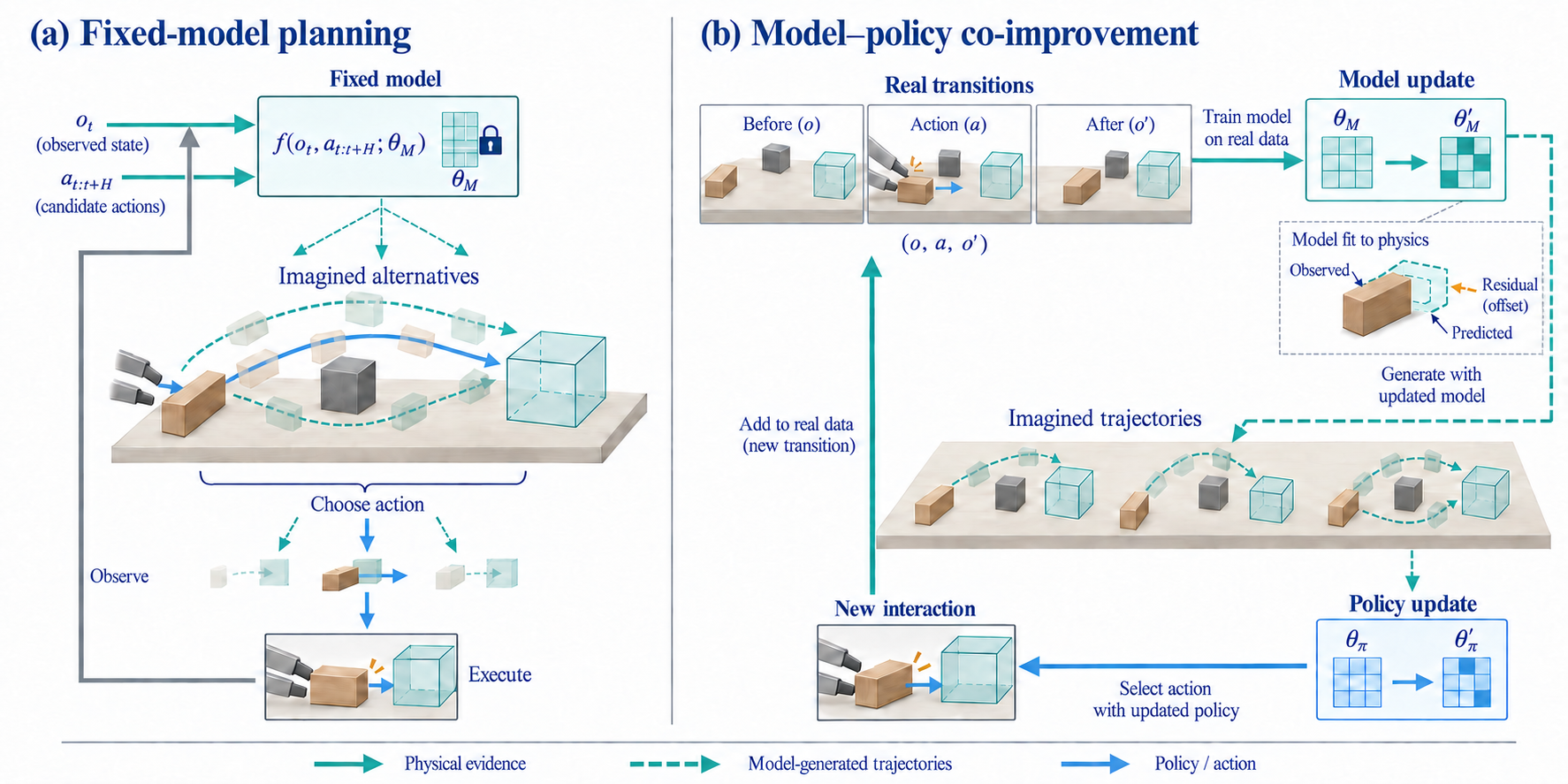}
\caption{Fixed-model planning and model--policy co-improvement. The fixed-model planner evaluates imagined action alternatives and executes a selected action while retaining model parameters. In the coupled loop, real transitions train the model, the updated model generates imagined trajectories for policy learning, and the updated policy collects further physical experience. Solid and dashed paths distinguish physical and model-generated experience. The observed--predicted offset is schematic; coupling alone does not establish inherited revision of the improvement procedure.}
\label{fig:worldroles}
\end{figure*}

Recent world-model methods further separate imagined data, planning, and learned control. WMPO optimizes a VLA using rollouts from a pixel-predictive model. RAW-Dream adds task-agnostic imagination, a vision--language reward source, and verification of unreliable imagined trajectories.\cite{REC_WMPO2025,REC_RAWDream2026} Imagine-RL instead keeps both the VLA and world model frozen while a learned controller or critic uses imagined visual and torque outcomes weighted by prediction confidence.\cite{REC_ImagineRL2026} MBDPO provides related algorithmic context by aligning policy optimization with trajectories searched in a latent world model.\cite{REC_MBDPO2026} The causal role of prediction differs across these designs: it supplies a training distribution, filters candidate experience, informs value estimates, or supports search. Figure~\ref{fig:worldroles} separates fixed-model use from model--policy co-improvement. The location of persistent change must be established before attributing a policy gain to improved modeling or a revised improver.

RISE and SILVR illustrate why the location of predictive learning matters. RISE separates future-observation prediction from progress or value estimation, using their outputs to train a policy in imagination. SILVR instead updates a visual planner from self-produced trajectories. In the former case, useful model changes must translate into better policy training; in the latter, they must improve the plans that execution subsequently attempts. A common video-quality metric cannot resolve either causal chain. The decomposition in RISE clarifies which component supplies transitions and which supplies evaluation, but it does not make their errors independent. A useful diagnostic distinguishes a mistaken predicted outcome from failure to execute a sound plan, since these failures require changes to different retained components. \cite{W001,W011}

VLAW and World-VLA-Loop feed new policy experience back into model learning, making the training distribution and predictive model mutually dependent. WoVR and VLA-MBPO address the reliability of model-generated rollouts, while VLA-RFT uses a world simulator with constructed reward signals. The shared difficulty is distributional: a stronger policy may visit states where the simulator is least reliable, and a policy optimized against that simulator may exploit precisely those errors. More realistic videos are therefore insufficient evidence of more useful learning experience. \cite{W006,W007,W017,W018,W019}

Coverage and supervision are as important as architecture. PlayWorld broadens dynamics experience through autonomous play; human-video dynamics transfer priors across embodiments; generative 3D worlds expand training environments. Driving-oriented reward world models further illustrate that the same vocabulary can hide different feedback and deployment assumptions. Comparisons should report how real evidence calibrates the model, which failures remain unsupported, and what physical transfer was actually tested. \cite{W020,W008,W021,W022}

The source of experience also determines which uncertainties a world model can resolve. Autonomous play supplies action-conditioned observations from the collecting robot, whereas human-video priors offer broader behavioral evidence with an embodiment gap. Generative environments can vary scenes extensively while retaining the assumptions of their construction process. These resources are complementary rather than interchangeable. The relevant comparison asks whether newly acquired evidence corrects an error that matters for action: an unseen contact mode, an unreliable object transition, or a gap in task coverage. More generated trajectories cannot by themselves compensate for a physical variable that the model never observes. \cite{W020,W008,W021}

Prediction can also guide the acquisition of further physical experience, connecting model learning to robot practice. EmbodiRSI collects corrections and targeted demonstrations in simulation aligned with the deployment scene, then fine-tunes in reality. RoboCoach uses a world model to guide subsequent skill learning. In addition to generating training trajectories, these approaches influence which states or skills the system practices. They bring experience allocation into the improvement process, whose revision and inheritance are examined in the next section. \cite{N_EMBODIRSI,N_ROBOCOACH}


\subsection{Memory and reusable skills}

Experience can also persist as explicit knowledge retrieved and invoked in later tasks. Compared with incorporating updates into model parameters, memory and skill libraries retain more direct units of experience, such as failure explanations, preconditions, and executable programs. Voyager and LRLL retain reusable program abstractions, LYRA converts human corrections into code, and ViReSkill revises and stores programs through visual failure diagnosis. A repair acquired in one task thereby becomes a capability available to later tasks while the foundation model can remain frozen. \cite{W118,W040,W033,W034}

As a skill library grows, maintaining existing skills becomes as important as discovering new ones. Uni-Skill uses available examples to fill capability gaps. EmbodiSkill distinguishes incorrect skill content from an executor failing to follow valid guidance before deciding how to revise it. PRACTICE trains a separate skill learner to edit the library, making the users and producers of skills separate objects of study. The former determine how a task is executed; the latter determine how new experience enters reusable capability. \cite{W035,W041,W042}

A reusable skill moves the burden of generalization from reproducing an entire trajectory to selecting and instantiating an abstraction. This can make a successful correction available to later tasks, but the abstraction must preserve the conditions under which the correction worked. LYRA's human corrections and ViReSkill's visually grounded diagnosis represent different sources of evidence for a retained program. EmbodiSkill adds another distinction: revising a valid skill is inappropriate when the failure comes from not following it. A useful comparison separates diagnosis, skill editing, retrieval, and execution. Otherwise, a larger library may receive credit for an improvement caused by a stronger planner, while a retrieval failure may be misread as incorrect stored knowledge. For physical tasks, the interface should preserve prerequisites and expected outcomes so that reuse can be checked against the current situation. \cite{W033,W034,W041}

The role of memory also depends on its temporal scope. RoboSkill concerns persistent skill artifacts, the main protocols of MemBodied and Zeva use within-episode information for adaptation, and AGM explicitly maintains verified task progress. Episode memory helps interpret the current situation, progress records support continued execution, and persistent skills serve later independent tasks. These uses explain which capability improves when more information is retained. \cite{N_ROBOSKILL,N_MEMBODIED,W050,W051}

Self-directed play and evidence-managed knowledge reveal two further forms of persistence. RATs selects practice tasks, executes and repairs code policies, and consolidates successful behavior into a skill library that can be frozen for downstream evaluation.\cite{REC_RATs2026} RoboHarn-Evo maintains linked task-level and action-level physical knowledge, revising applicability and retrieval eligibility from execution evidence while retaining a fixed task--action schema and fixed execution primitives.\cite{REC_RoboHarnEvo2026} In both cases, later behavior can change through what is retrieved or reused. Their different retention units suggest different tests: library ablations isolate reusable skills, whereas frozen knowledge snapshots isolate the effect of accumulated applicability information. Those tests should preserve the surrounding executor so that a knowledge gain is not confounded with an unreported controller change.

\subsection{Executable programs and agent harnesses}

Knowledge in a skill library becomes useful through the system that executes it. A harness organizes the tool interfaces, context construction, skill routing, recovery routines, and execution checks around a model. Revising it changes how existing models and skills work together. ASPIRE retains validated program repairs, SHAPER evolves skills and planning context, and Zetta revises runtime assets around a frozen policy. Harness VLA and HarnessPAI organize reusable skills and action interfaces. These approaches extend persistent improvement to system orchestration. \cite{W027,W029,W030,W031,N_HARNESSPAI}

Graph-as-Policy makes editable control structure explicit, while PhyAgentOS and skill-graph assembly systems emphasize session boundaries, verification, and reusable interfaces. VASO adds formal checks over specified skill contracts; CRISP and SERP illustrate correction and replanning whose persistence must be established separately. CaP-X supplies useful infrastructure for studying coding agents, but the availability of an editing interface alone establishes neither retained learning nor recursion. \cite{W036,W038,W045,W043,W046,W044,W039}

Executable artifacts make some failure causes easier to locate than a monolithic parameter update. A repaired branch can change how a task recovers, and an edited graph can change which skill follows an observation. Yet editability creates dependencies: a new tool interface can invalidate an older skill, while a routing change can alter which policy failures are exposed. ASPIRE's retained repairs and graph-based control structures should therefore be analyzed at the artifact level as well as through end-task success. The key persistence test is whether an accepted change remains available in a fresh attempt with its required dependencies intact. Only when that artifact also changes how later updates are produced does the analysis move from this section's learner improvements to process revision. \cite{W027,W036,W045}

HarnessPAI illustrates two coupled but distinguishable updates: executable programs remain fixed within a rollout and evolve between rollouts, while successful trajectories can subsequently train the wrapped action model. Its stored repair knowledge supports reuse across tasks. These paths motivate separate ablations for program improvement and model post-training: a final system gain does not identify their individual contributions, and persistence of an execution program does not by itself establish revision of its updater. \cite{N_HARNESSPAI}

RoboFoundry and Physical Coding extend persistent artifacts to the organization of the execution system. Compared with repairing one skill, a system-level edit can affect context, tool invocation, and task recovery together. Policies, models, memory, and programs consequently form an interdependent capability structure in which experience can change execution through different entry points. A retained program that also generates and selects later updates simultaneously serves as part of the improver. \cite{N_ROBOFOUNDRY,N_PHYSICALCODING}

RegenHarness makes a useful distinction between a proposed modification, verified physical progress, and a released system version. Its design retains changes to context, templates, routing, and recovery across missions, with regression checks and versioned rollback.\cite{REC_RegenHarness2026} This explicit release boundary makes the edited artifact and its later use inspectable. The deployment cases and the paper's proposed controlled RSI protocol nevertheless answer different questions: the former illustrate operation and evidence gating; the latter specifies how a revised improver should be tested. This distinction connects executable persistence to the inherited process revisions examined next.

\section{From Self-Improvement to Recursive Improvement}
\label{sec:recursion}

\begin{figure*}[t]
\centering\includegraphics[width=\textwidth]{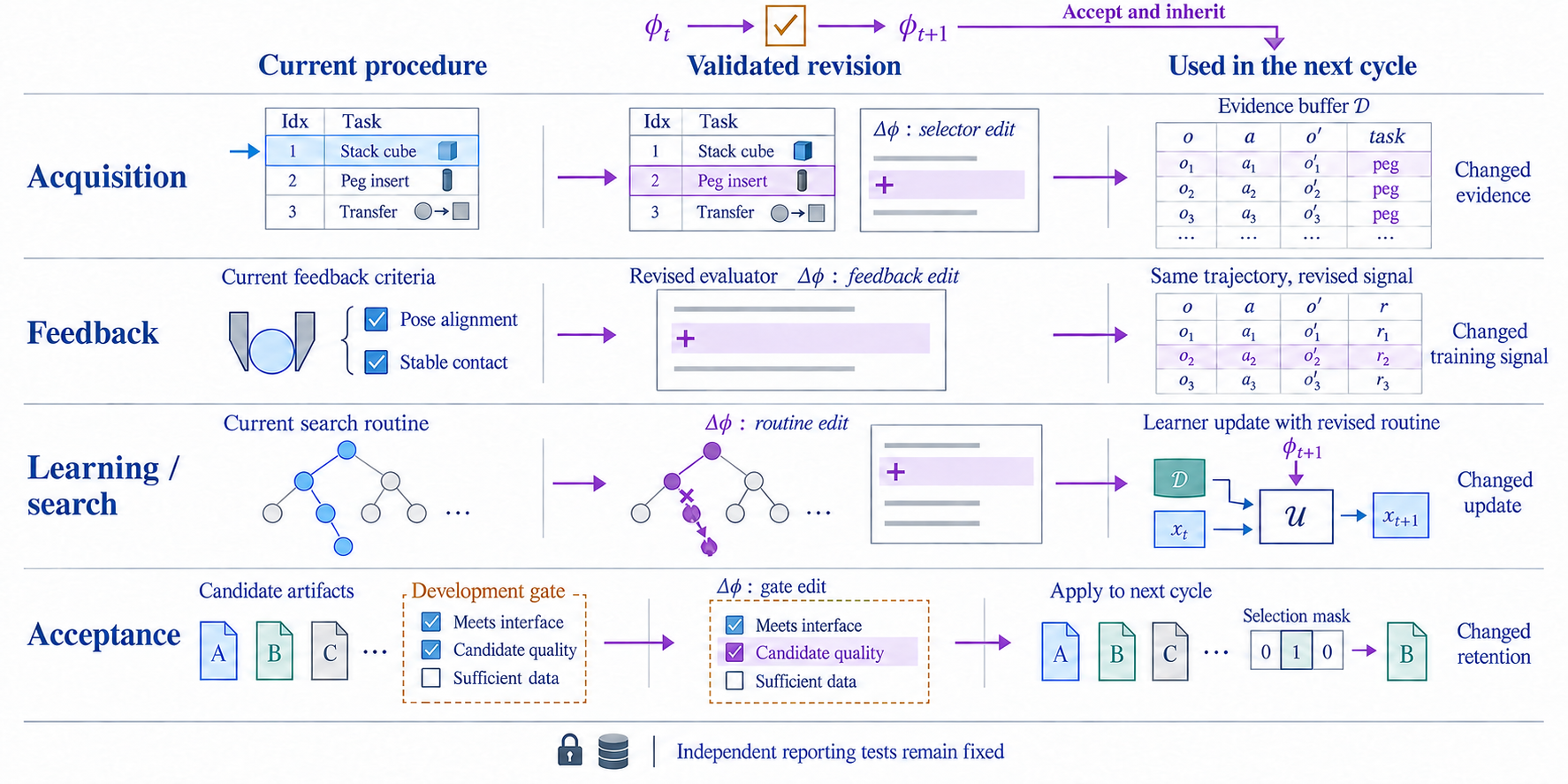}
\caption{Four locations for inherited improver revision. Changes to acquisition alter the selected experience; changes to feedback alter the training signal; changes to learning or search alter the update routine; changes to a development gate alter which candidates are retained. Highlighted differences are accepted, inherited, and used in a later cycle. The rows describe alternative intervention locations. Development-gate revisions remain subject to declared external constraints, while independent reporting tests stay fixed for the comparison.}
\label{fig:targets}
\end{figure*}

The preceding mechanisms explain how experience changes and preserves task capability; the learning procedures that produce these changes can themselves become objects of revision. When a system uses operational results to revise experience acquisition, feedback construction, or learning code, then reuses the revision in later rounds, experience also changes how capability is produced. Figure~\ref{fig:targets} locates the main revision points. Understanding these approaches requires tracing both the modified component and its inheritance: acquisition and curricula control what is encountered next, rewards and evaluators determine learning signals, and training and research programs organize updates. What is retained and how it enters subsequent learning determine the scope of recursion.

\subsection{Revising experience acquisition and curricula}

Curriculum methods extend adaptation to the distribution of training experience while the student policy continues to learn.\cite{liang2025structured} Teacher–student learning selects tasks using learning progress; reverse curricula expand feasible starting states from the goal, and Goal GAN generates goals of intermediate difficulty. PAIRED and ACCEL extend this control to environment generation and editing using regret-related signals. These mechanisms can improve later policy updates by changing which evidence is encountered. However, a changing curriculum does not by itself imply that the curriculum-generating procedure has been revised. This paper separates the selected tasks, the selector's retained state, and any inherited modification to its selection rule.\cite{F_TSCL,F_REVERSE,F_GOALGAN,F_PAIRED,F_ACCEL}

Curriculum and environment generation extend this control to the tasks a learner encounters. POET, EnvGen, and OMNI-EPIC explore different forms of task or environment generation, while H-GRAIL relates learning progress to hierarchical goals. A learner can thus practice on an evolving problem distribution instead of relying entirely on a predefined task list. Two changes are possible: a generator can produce new tasks under its existing rules, or the system can revise those rules for later practice. The former expands experience; the latter brings acquisition itself into recursion. \cite{W076,W075,W077,W073}

Curriculum construction exposes the difference between changing the problems encountered and changing the process that constructs them. POET retains environment--policy pairs and uses transfer between them, allowing a solution useful in one environment to become a stepping stone elsewhere. The evolving collection is an important learning resource, but an expanding archive alone does not show that its evolutionary procedure has changed. A fixed generator can propose increasingly difficult environments because its inputs change. A process-revision claim must identify what was inherited beyond those outputs, such as a revised acquisition policy or executable generator, and show how it governed a subsequent collection round. Immediate success is an ambiguous signal: easier sampling can raise it without broadening competence, whereas harder practice can lower it while making later learning more useful. \cite{W076,W075,W077}

SimWorld Studio makes executable environment construction and repair part of the learning infrastructure. RECALL directs demonstrations toward uncertain or failed states, while SAGE links active acquisition to intervention in driving. SEES connects failure analysis to targeted skill development. These systems reveal a recurrent design tension: concentrating on current weaknesses can improve sample usefulness while narrowing coverage or increasing reliance on a particular teacher. Evaluation must measure retained broad competence, acquisition cost, and the origin of corrective supervision. \cite{W074,W063,W024,N_SEES}

Failure-driven acquisition links deployment to the next training distribution. F4R diagnoses deployment failures, reconstructs corresponding interactive scenes, refines the policy through sim--real training, and redeploys it to collect further failures. Its initial smartphone capture and workspace alignment involve human setup; subsequent refinement is organized as an automated loop.\cite{REC_F4R2026} TwinRL uses a reconstructed digital twin to warm up learning and select informative configurations for real, human-assisted rollouts.\cite{REC_TwinRL2026} POEL steers self-generated goals toward a user-specified purpose, while REAL combines exploration, intent clarification, and online learning for mobile manipulation.\cite{REC_POEL2025,REC_REAL2026} These methods differ in how they make new evidence accessible: reconstruct a failure, prioritize a configuration, constrain exploration by purpose, or actively inspect an open environment. The next comparison is whether the selector simply executes its original rule or inherits a revised procedure for deciding what to acquire.

\subsection{Revising rewards and evaluators}

Feedback methods differ in who supplies the task criterion and how that criterion becomes an optimization signal. Preference learning fits rewards to human comparisons; PEBBLE combines preference feedback with unsupervised exploration and replay relabeling. RoboCLIP uses a video–language model and a demonstration to construct rewards, while RL-VLM-F derives comparisons from a vision–language model. Vision-language rewards further examine transferring semantic judgment into reinforcement learning. This shifts the bottleneck from manually programming every reward to maintaining a valid link between the proxy and physical success. When the reward model evolves, retained evaluator changes and independently assessed physical outcomes must be recorded separately.\cite{F_PREFRL,F_PEBBLE,F_ROBOCLIP,F_RLVLMF,F_VLMREWARDS}

Task generation and feedback generation often need to be designed together. Agentic Skill Discovery connects task proposals, reward generation, skill training, and retention, with an additional visual check of task completion. A new task supplies a useful skill-learning signal when its objective and feedback agree. Acquisition, evaluation, and training therefore become interdependent parts of the process. \cite{W032}

Reward code provides a concrete carrier for process revision. Eureka generates and revises executable rewards from training feedback, changing the optimization signal used by later policies. DrEureka extends automated design to training choices relevant to transfer, while reflective task adaptation connects failure diagnosis to rewards and curricula. A revised reward retained for the next training round creates a local feedback path from learning outcomes to learning conditions. The policy objective or training conditions change while the outer design procedure can remain fixed. \cite{W052,W086,W015}

Reward revision changes the objective seen by a downstream learner, so its effect cannot be reduced to a better terminal score under the revised reward itself. Eureka's executable reward programs make this intervention inspectable: a changed term can alter which intermediate behavior the optimizer favors. DrEureka broadens the design problem to transfer-related choices, where favorable simulated learning still requires a hardware check. The comparison should hold the intended task outcome conceptually separate from the training signal and record which constraints remained fixed. If a reward change merely redefines failure as success, internal optimization can improve without physical progress. If the change supplies more informative feedback for the same intended outcome, it may improve subsequent learning even though the outer reward-generation procedure remains unchanged. \cite{W052,W086}

Learned rewards provide another representation of feedback. RoboReward, Robo-Dopamine, Robo-Dopamine 2.0, and large-scale reward models improve feedback in different ways, while VERITAS connects execution-time verification to the retention of data for later training. Even a reward model frozen during downstream learning can redirect policy updates. When the reward model continues to absorb experience, policy and feedback become coupled: the policy determines which trajectories the evaluator observes, and the evaluator influences how the next policy is trained. \cite{W053,W054,W055,W056,W057}

Feedback adaptation now spans the construction of examples, reward fitting, and reward reuse. DenseReward synthesizes failure trajectories to broaden the coverage of a language-conditioned progress model. RAPID combines parallel simulation, batched vision--language preferences, and reward stabilization to reduce the cost of obtaining learning signals.\cite{REC_DenseReward2026,REC_RAPID2026} eVTA0 learns dense success-probability rewards from mixed-quality policy rollouts and updates the reward model as reinforcement learning collects further data.\cite{REC_eVTA02026} This creates a concrete pathway through which policy-dependent evidence changes future feedback. BEE takes a complementary approach to corrective supervision: it learns the consistency of human interventions and uses that information to constrain reinforcement learning around a frozen VLA.\cite{REC_BEE2026} The comparison must therefore distinguish who supplies the criterion, how feedback is estimated, and when an accepted feedback change becomes part of later training. Holding the rollout set fixed can isolate feedback use; allowing new rollouts measures the coupled acquisition and learning process.

Jointly evolving an actor and its evaluator creates an attribution problem: their internal agreement may rise while physical validity falls. The Red Queen Gödel Machine offers a digital example of changing evaluators while maintaining external anchors and version boundaries. For physical AI, reward calibration should be checked against measurements or labels unavailable to the updating system. Independence is a relationship among data, models, and decisions, not merely the use of a differently named judge. \cite{W112}

\subsection{Revising learning and research procedures}

When training and research programs become editable, a system can change the method used to produce updates. ENPIRE allows an agent to modify training code and policy-learning choices, linking execution logs to subsequent policy development. Compared with repeated fine-tuning under a fixed algorithm, this exposes algorithm selection and implementation to revision. Robot tasks, safety interfaces, and reset conditions constrain the experiment, while the revised program organizes the next learning round within those conditions. \cite{W028}

A program can be both the artifact being improved and the tool that produces further artifacts, a relationship especially direct in digital-agent research. STOP edits an optimization program that invokes a language model. Gödel Agent and Darwin Gödel Machine expose agent code to modification and evaluation. Dream-RSI revises exploration procedures using replayable search experience, while Escher-Loop separates task and optimizer agents. These methods offer references for constructing, retaining, and rerunning improvement programs. Robotic applications additionally connect them to physical experience, action interfaces, and experimental constraints. \cite{W109,W110,W111,W119,W114}

The correspondence between task competence and improvement competence is especially important when importing digital-agent ideas into robotics. In STOP, executable code is both an output of optimization and part of the machinery that performs it. A manipulation policy does not generally have that correspondence: grasping more reliably does not give the policy an interface for editing a training script. ENPIRE supplies such an interface through agent-managed policy development, making the causal connection explicit at the system level. The appropriate comparison is between complete experimental procedures, including their access to logs, code, and evaluation, rather than final policies alone. A procedure that benefits from additional human debugging or a different foundation model has changed under more than the claimed autonomous revision. \cite{W109,W028}

Physical design broadens the object of improvement further. Auto-Robotist retains knowledge used to propose later morphologies; RoboMoRe couples morphology and reward design; graph-based co-design and SwarmCoDe study changes to bodies or collective structure. A fluid-control scientific agent co-evolves design agents and controllers in simulation. These examples prevent a policy-only definition of physical improvement, but simulation-based design, fabricated hardware, and autonomous physical redesign must remain distinct evidence categories. \cite{W082,W083,W084,W085,W078}

\subsection{Inheritance and the boundary of a recursive claim}

Revisions to acquisition, feedback, and training code converge on a common relationship: accepted changes remain active in later rounds. Connecting the parent program, the experience that motivated a revision, the accepted version, and its next update reconstructs this inheritance. AgenticRobotics and harness-engineering work provide references for organizing execution and version information. These records distinguish a temporary repair from a reusable process change in an implemented system. \cite{W122,W091}

Inheritance must also preserve the evidence needed to interpret a revision. A successful final artifact does not reveal whether an earlier rejected candidate supplied the idea, whether a human repaired the environment, or whether the accepted procedure was ever used again. An audit should connect proposals and failures to the accepted version and then to the next learner update. This is more informative than counting self-editing calls, since many edits may be inert or confined to current-task execution. Where a world model or skill library serves both execution and training, the record should identify which role changed. The same stored artifact can support persistent capability without yet establishing a revised improvement process.

RoboFoundry explicitly separates task-level repairs from changes promoted to a persistent general system. General edits are tested through replay of other tasks' pre-repair traces, using the model-based system as verifier of proposed repairs. This makes the reuse scope inspectable, while leaving development verification distinct from an independent physical audit. This paper therefore asks whether a retained edit improves future proposal generation, task execution, or both, and whether that distinction is isolated by the reported comparison. \cite{N_ROBOFOUNDRY}

PhysicalRSI 1.0 combines a frozen foundation model with an editable harness, using candidate generation, selection, and inheritance for persistent harness revision. The key analytical object is the function of an inherited edit: whether it changes task execution, subsequent learning, or both. Final-system leaderboard performance measures task capability; improver gain requires a comparison of the subsequent learning gains produced by the revised procedure. General Robotics AutoEngineering further broadens the engineering scope of editable robot systems. \cite{N_PHYSICALRSI,W130}

\section{Physical Conditions for Sustained Improvement}
\label{sec:physical}

\begin{figure*}[t]
\centering\includegraphics[width=\textwidth]{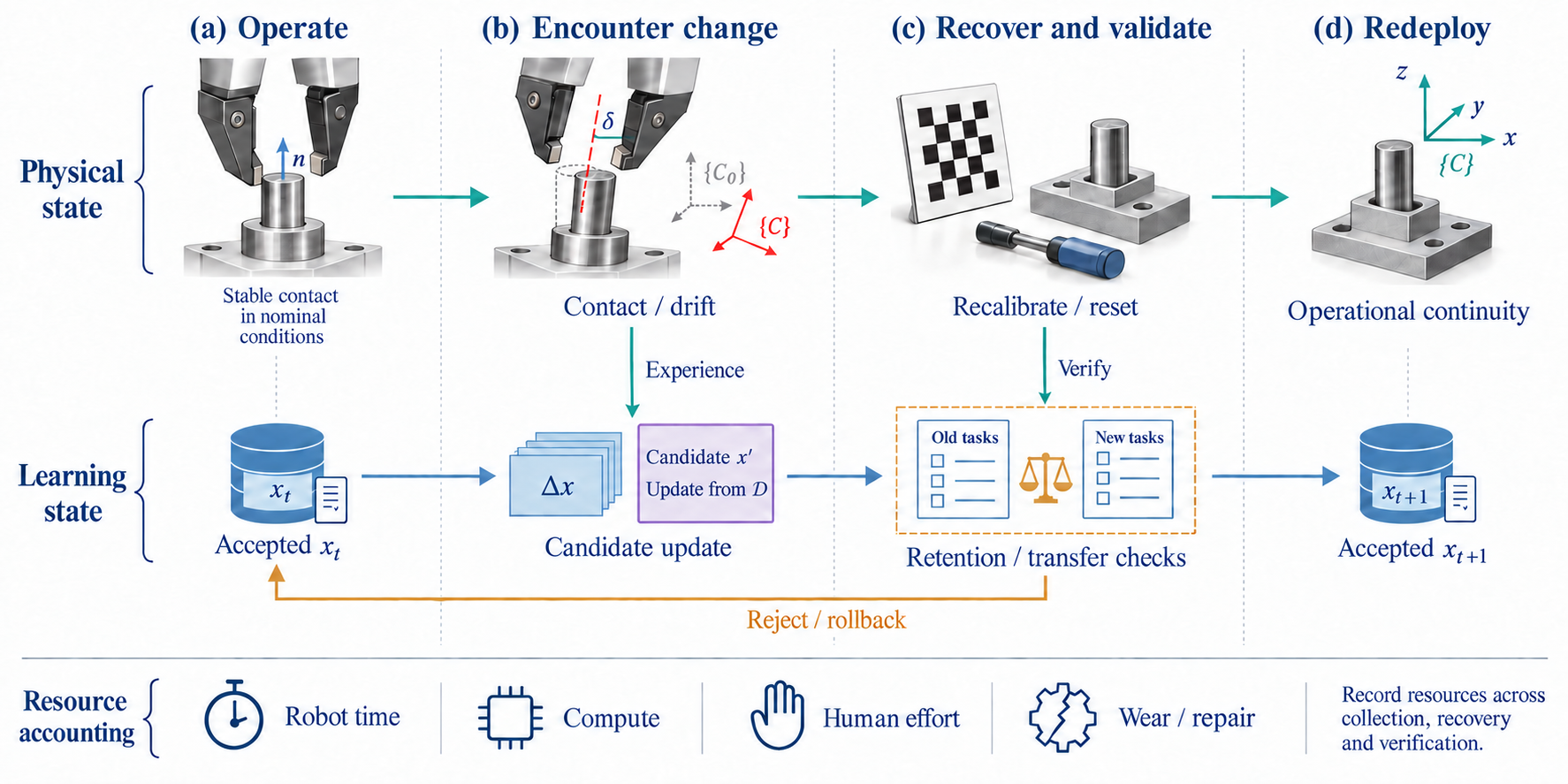}
\caption{Physical continuity and learning-state retention are separate, coordinated processes. Contact changes or calibration drift can require physical recovery before redeployment. Candidate changes to retained state are checked on earlier and new tasks, then accepted or rolled back to the previous checkpoint. Recovery is not itself a learning update. The resource ledger spans collection, recovery, updating, and verification, including robot time, computation, human effort, and wear or repair.}
\label{fig:continuity}
\end{figure*}

Accumulating long-term progress over successive updates requires a robot to keep obtaining valid experience and remain operational after each change. An action that appears visually plausible may fail at contact, an unrecoverable operation may interrupt data collection, and a new skill may lose its usefulness under environmental change or further learning. These conditions extend across experience acquisition, update validation, and redeployment (Figure~\ref{fig:continuity}). Contact and model validity affect the reliability of feedback; recovery determines whether interaction can continue; and retention determines whether gains accumulate. Update control and full costs further constrain which changes the system can attempt and how long improvement can continue.

\subsection{Contact and model validity}

Physical feedback is constrained by what can be sensed and safely attempted. Visually plausible grasping need not establish force closure, secure contact, or successful insertion. Model error can therefore affect both action selection and the labels used to accept an update. Tests should distinguish predicted success, observed completion, and independently verified task outcomes. A model used for training should be evaluated on failures and interventions, not only on trajectories similar to its successful demonstrations. \cite{W006,W017,W020,W080}

Contact tasks make the dependence between prediction and feedback especially visible. A grasp image can show geometry without revealing slip, and apparent alignment in assembly can coexist with insufficient insertion depth. A world model and reward model that rely on the same visual cues may therefore accept the same physical error. Object-state, force, or tactile measurements chosen for the failure mechanism can test conditions that visual feedback misses. The relevant criterion is whether the measurement distinguishes the outcome the task actually requires.

Simulation can make experimentation reproducible, but it does not remove the transfer obligation. DrEureka, generative-world training, and TTT-Parkour illustrate different ways to use simulation for physical policy improvement. The relevant evidence chain records what was reconstructed or randomized, which parameters were calibrated using real observations, and what was subsequently checked on hardware. Simulation and real-world budgets should be reported separately before any aggregate efficiency claim is made. \cite{W086,W021,W069}

Sim-to-real transfer and rapid adaptation clarify why robustness should be separated from persistent self-improvement. Visual and dynamics randomization diversify training conditions before deployment; dexterous in-hand manipulation demonstrates this strategy in a physical control setting. RMA infers deployment conditions online through an adaptation module, and its manipulator-arm extension applies rapid adaptation to another embodiment. Such mechanisms can change behavior effectively without inheriting a revised learning procedure. For the survey, the relevant questions are whether the deployment update persists after reset, whether later learning changes because of that update, and whether any claimed gain transfers under independently varied physical conditions.\cite{F_DOMAINRAND,F_DYNAMRAND,F_RMA,F_DACTYL,F_RMAARMS}

\subsection{Recovery and continuity}

Learning stops before the next sample can be collected when a robot drops an object, leaves its workspace, or enters a state from which the task cannot continue. Recovery returns such states to usable interaction conditions. MEDAL++ and RoboFuME incorporate recovery into autonomous practice, RoboClaw pairs forward operation with recovery, and RADAR coordinates collection, filtering, and resets. These mechanisms improve access to continued experimentation, reflected in autonomous attempts, recovery success, human intervention, and task performance. \cite{W070,W071,W037,W062}

Autonomous improvement depends on keeping the robot in states from which further informative interaction remains possible. Leave no Trace learns a reset policy; multi-task reset-free learning uses relationships among tasks to continue practice without repeatedly restoring a single initial state. The Ingredients of Real-World Robotic Reinforcement Learning combines reward acquisition, policy learning, and autonomous practice, while ARIEL uses prior data to reduce the burden of autonomous task learning and sequencing. MoReFree studies reset-free learning with world models. These works motivate treating recovery as part of evidence acquisition: two systems with the same nominal rollout count may incur different intervention, reset, and inaccessible-state costs.\cite{F_LEAVENOTRACE,F_MTRF,F_ARIEL,F_R3L,F_MOREFREE}

Recovery designs differ in their dependence on people and environmental preparation. A robot can reduce manual resets while still using a prepared workspace, a backup controller, or expert corrections. Reset-free agile-driving work illustrates the role of control and exploration conditions, while RECALL retains expert recovery supervision. Recovery also changes the states accessible in the next round, affecting both operating costs and the distribution of learning data. \cite{W064,W063}

Recovery also determines what the next experiment can teach. Restoring only convenient states can produce a stable stream of trials while excluding precisely the failures a policy needs to learn from. Conversely, treating every abnormal state as another exploration opportunity can make restoration slower or less reliable. The forward--recovery organization in RoboClaw and the active recovery supervision in RECALL address different parts of this continuity problem. Their contribution should be described through the states they make available and the assistance needed to reach them. Human participation is heterogeneous: demonstrations, evaluator labels, timed action corrections, object restoration, and safety monitoring consume different kinds of effort. Eliminating one does not establish the absence of the others, and their costs scale differently as the robot fleet or task repertoire grows. \cite{W037,W063}

Execution-time recovery can preserve access to learning even when it does not itself create a persistent update. AgentChord anticipates failures in a task graph and monitors when to trigger a recovery branch, while Physical Agentic AI uses typed skills and per-dispatch checks to constrain multi-robot execution.\cite{REC_AgentChord2026,REC_PhysicalAgentic2026} Such mechanisms change the failures a learner encounters and the interventions needed to continue. Evaluations should therefore record both attempted learning updates and the recovery machinery surrounding them, including fixed fallback behavior, human resets, and interventions that prevent an attempted action from reaching the robot.

\subsection{Retention and distribution change}

Once a robot can keep operating, long-term benefit depends on whether new and existing capabilities coexist. CRL-VLA and LifeLong-RFT study retention during sequential updates, whereas Natural Continual Learners demonstrates strong simple baselines under its conditions. These outcomes reflect the joint effects of task overlap, representation, replay, and training objectives. Real-world continual VLA experiments provide complementary settings: externally supplied task and demonstration sequences support retention analysis, while autonomous experience acquisition forms a separate dimension. \cite{W066,W067,W065,N_CONTINUALVLA}

Sustained improvement has two distinct failure modes: losing earlier capabilities and losing the capacity to acquire new ones. EWC constrains changes to important parameters, while GEM uses stored examples to constrain updates that interfere with previous tasks. Continual World makes retention and forward transfer explicit evaluation targets in robotic task sequences. Studies of plasticity loss show why preserving old behavior alone is insufficient: repeated learning can leave a network increasingly difficult to train. The survey should therefore report old-task retention and new-task learning separately. An apparent stable system may simply have stopped adapting, whereas rapid new-task improvement may conceal accumulated forgetting.\cite{F_EWC,F_GEM,F_CONTINUALWORLD,F_LOP_RL,F_LOP_NATURE}

Forgetting also affects executable artifacts. A new skill may conflict with an old interface, increase retrieval ambiguity, or encode a local shortcut. Library Drift and SkillFlow show why library growth should be evaluated through use and downstream benefit rather than artifact counts. Scientific Amnesia further distinguishes preserving task behavior from retaining knowledge about how to train the next system. These digital findings motivate physical regression tests, while their quantitative results remain specific to their original environments. \cite{W124,W121,W125}

Retention must cover the whole deployed system. A frozen policy can still lose an effective behavior if a new selector stops choosing it, a memory entry overrides its context, or a revised skill interface no longer calls it correctly. An unchanged library may also become unreliable after calibration or material properties drift. Rollback is therefore more demanding than restoring one checkpoint: compatible model, memory, tool, and controller versions may need to be restored together. VLA retention studies and studies of skill-library drift motivate complementary checks, since parameter stability and artifact availability measure different things. A sustained loop needs to distinguish forgotten competence, inaccessible competence, and competence invalidated by changed physical conditions before choosing whether to retrain, repair retrieval, or recalibrate. \cite{W066,W067,W124,W023}

Hardware wear, changed calibration, and environmental drift make retained artifacts conditional. A formerly useful grasp parameter or reward threshold can become incorrect without any software edit. World-model-feedback adaptation studies the triggering and stopping of learning under change; SAGE connects risk to intervention. Sustained improvement requires a policy for detecting invalid knowledge, revalidating it, and distinguishing adaptation to drift from learning a transferable new capability. \cite{W023,W024}

Recent VLA studies show why forgetting must be interpreted through the training protocol. Memory Anchors prioritizes replay examples where conflicting tasks share representations, aiming to preserve behavior at points of interference. A separate study of pretrained VLAs reports resistance to forgetting under its continual-learning and replay conditions, including recovery of degraded skills.\cite{REC_MemoryAnchors2026,REC_ResistantForgetting2026} These observations concern different task conflicts, pretrained representations, and replay supplies. They motivate reporting the sequence of tasks, the retained data, performance before and after relearning, and the cost of recovery. A final average can conceal both transient regressions and unequal practice on earlier tasks.

\subsection{Update control and full costs}

Accepting a new version also requires considering its effect on real-time execution. Additional perception, planning, or verification changes latency and computational competition, so a component that performs better offline can behave differently in deployment. VASO checks specified execution conditions through declared skill contracts, while staged deployment and rollback control the operational effects of revisions. Formal checks apply within their abstractions and assumptions; contact and sensing conditions also require validation during execution. \cite{W043,W091,W060}

Update control introduces a trade-off between protection and visibility. A fixed fallback controller can keep an experimental policy within a restricted operating envelope, yet frequent fallback may conceal deterioration of the policy being improved. The deployed claim then concerns the combined system, and intervention records are necessary to attribute performance. Formal checks over skill contracts, such as those considered by VASO, provide a different kind of evidence from a favorable learned reward: their scope depends on the stated abstraction and assumptions. Neither should be generalized to unmodeled sensor or contact failures. Preserving these boundaries keeps failed updates identifiable and reversible, rather than letting recovery infrastructure make every candidate appear equally successful. \cite{W043}

The learning objective also interacts with execution timing and motion quality. ARLI accounts for actions already committed during slow policy inference, so the learner sees the consequences of asynchronous execution. SmoothVLA combines task-success reward with a trajectory-jerk term, bringing motion smoothness into post-training.\cite{REC_ARLI2026,REC_SmoothVLA2026} Social DSRL studies learned diffusion steering for navigation under a performance-preserving adaptation objective, with a hardware-in-the-loop setting in the inspected evidence.\cite{REC_SocialDSRL2026} These cases show why physical improvement is not exhausted by goal completion: timing, actuator-friendly trajectories, and acceptable behavior during adaptation determine whether a policy update can be used in continued operation.

Improvement cost is a vector rather than a single token count: real robot time, simulated interaction, compute, teacher demonstrations, human maintenance, resets, failed trials, and hardware wear can trade off. Reporting only successful retained trajectories omits the acquisition attempts that produced them. Likewise, counting only the final training run omits search and discarded candidates. A claimed gain in sample efficiency is interpretable only relative to the resources it reduces and those it increases.

\begin{figure*}[t]
\centering\includegraphics[width=\textwidth]{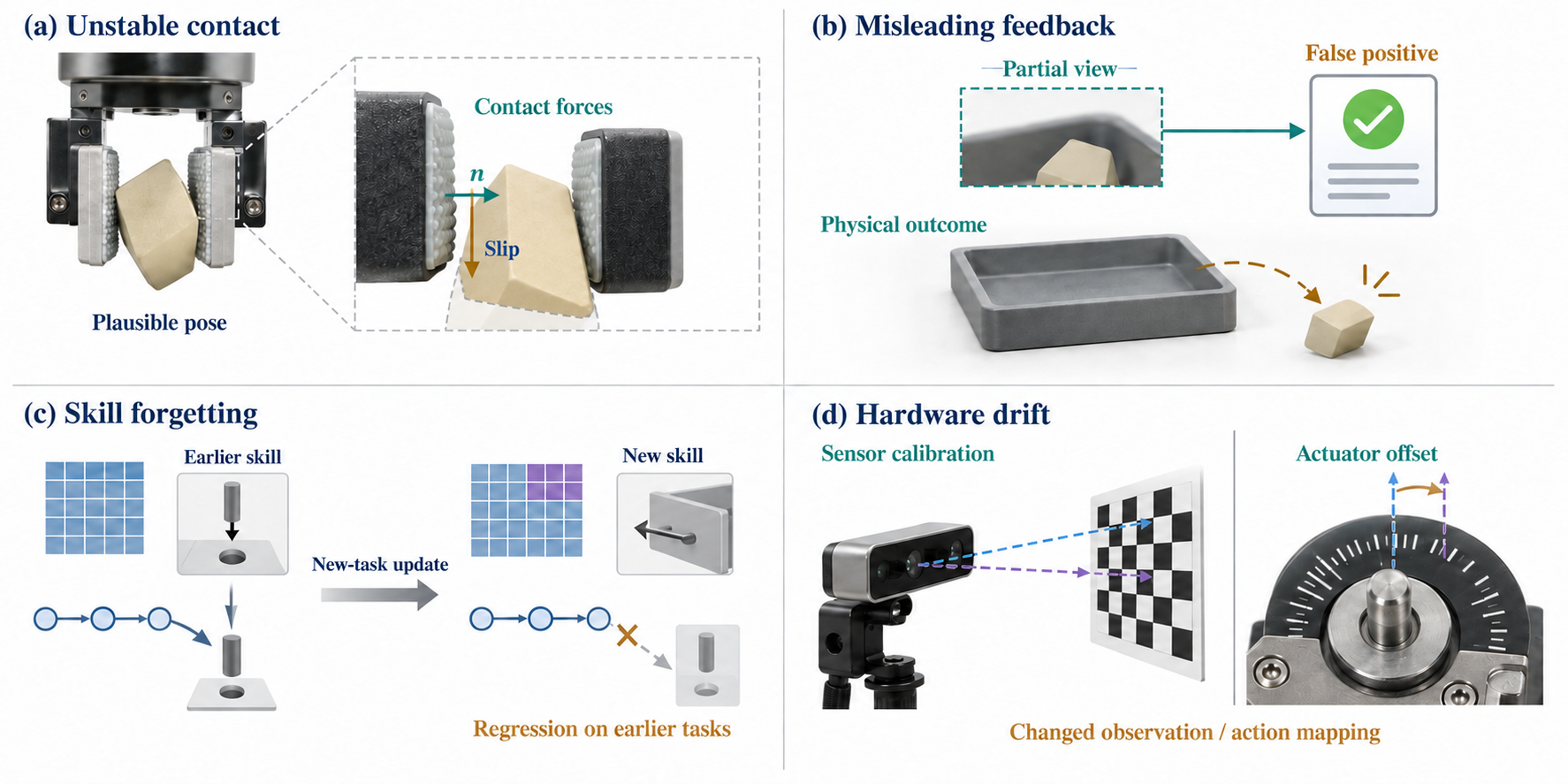}
\caption{Conceptual physical failure modes. (a) A plausible pose can yield unstable contact, illustrated by contact-normal and tangential-slip directions. (b) Partial observations can produce a false-positive judgment despite an unsuccessful physical outcome. (c) A new-task update can disrupt an earlier skill. (d) Sensor calibration or actuator offsets can change the observation--action mapping. The scenes, state changes, and offsets are illustrative rather than experimental measurements.}
\label{fig:failure}
\end{figure*}

These factors explain the distance between a single update and long-term benefit. Incorrect feedback directs learning toward ineffective behavior, recovery failures reduce available experience, and forgetting or hardware drift erodes acquired capability. Figure~\ref{fig:failure} summarizes these distinct failure paths. Evaluating sustained improvement therefore involves both task performance and the conditions that keep further learning possible.

\section{Evidence and Evaluation}
\label{sec:evaluation}

Mechanism analysis explains how a system changes; experimental results establish what those changes achieve. A similar increase in task success may result from a better policy, reusable skills, additional supervision, or a revised learning procedure, so interpretation depends on the evaluated object, resource inputs, and evaluation boundary. We first establish common dimensions for comparing methods, then examine benchmark designs and quantitative results, before considering controlled comparisons between reference and revised improvers from shared learner starts. This connects task capability, inherited process revisions, and subsequent improvement capacity to distinct, testable claims.

\subsection{Common dimensions for method comparison}

Comparing methods requires identifying the updated object, the reset boundary beyond which changes persist, and the states used in subsequent tests. Feedback sources, human assistance, experimental settings, and resource inputs then establish the conditions of the comparison. These fields support comparisons between related mechanisms and clarify which claims different configurations of the same method can support. Table~\ref{tab:evidence-fields} specifies the information needed to make these comparisons alongside the L0--L3 evidence levels.

\begingroup\small
\setlength{\tabcolsep}{3pt}
\begin{table}[tb]
\centering\fontsize{8.5}{10.5}\selectfont
\caption{Evidence fields used independently of the L0--L3 rubric}
\label{tab:evidence-fields}
\begin{tabular}{>{\raggedright\arraybackslash}p{0.28\columnwidth}>{\raggedright\arraybackslash}p{0.64\columnwidth}}
\toprule
\textbf{Field} & \textbf{What must be recorded}\\\midrule

Persistent change & Weights, world model, memory, skill code, harness, reward, or update procedure; record functional role.\\[3pt]
Feedback and setting & Physical measurements, human labels, learned judge, simulator, or hybrid; distinguish physical and digital evidence.\\[3pt]
Retention boundary & Across attempts, independent episodes, tasks, or deployments; specify every reset.\\[3pt]
Human and resource input & Demonstrations, resets, approval, maintenance, corrections, compute, time, and discarded trials.\\[3pt]
Claim and control & Capability gain; inherited process revision; improver gain; shared starts, fixed references, and evaluator independence.\\[3pt]
\bottomrule
\end{tabular}
\end{table}
\endgroup

This organization reveals substantive differences between methods with similar names. Training a new policy, retaining an execution repair, and revising a training program can all improve task performance, but they preserve different objects for different later uses. Table~\ref{tab:assessments} places representative methods alongside one another by these comparison objects, making connections across the technical approaches easier to trace.

\begin{table*}[tp]
\centering\scriptsize
\setlength{\tabcolsep}{4pt}
\caption{Improvement objects and evaluation focus of representative methods}
\label{tab:assessments}
\begin{tabular}{P{0.22\textwidth}P{0.24\textwidth}P{0.49\textwidth}}
\toprule
\textbf{Method} & \textbf{Comparison object} & \textbf{Evaluation focus}\\\midrule
Zeva \cite{W050} & Task context & Memory boundaries across episode resets.\\[2pt]
MemBodied \cite{N_MEMBODIED} & Episode memory & Within-episode memory and persistent benefit across episodes.\\[2pt]
RECAP \cite{W003} & Policy updates & Retained capability gains under a fixed learning procedure.\\[2pt]
Learning While Deploying \cite{W004} & Fleet policy learning & Effects of fleet experience and model versions.\\[2pt]
RISE \cite{W001} & Model-assisted policy learning & Training gains and the contribution of planning.\\[2pt]
Motus2 \cite{W002} & Policy and planning & Policy retention and planning gains under shared initialization.\\[2pt]
VLAW \cite{W006} & World model and policy & Role of retained models in subsequent policy training.\\[2pt]
ASPIRE \cite{W027} & Reusable skills & Skill retention and the skill-improvement process.\\[2pt]
SHAPER \cite{W029} & Skill harness & Changes to execution logic and learning procedures.\\[2pt]
Zetta \cite{W030} & Runtime assets and skills & Reuse of retained assets in later tasks.\\[2pt]
HarnessPAI \cite{N_HARNESSPAI} & Programs and harnesses & Execution-program evolution, model post-training, and updater revision.\\[2pt]
RoboFoundry \cite{N_ROBOFOUNDRY} & Task repairs and general rules & Scope of rule inheritance and independent evaluation of later gains.\\[2pt]
FIND \cite{N_FIND} & Training support and policy & Contribution of training support to the evaluated policy.\\[2pt]
SkillSpace Shooting \cite{N_SKILLSHOOT} & Policy and test-time search & Contributions of retained policy changes and search.\\[2pt]
EmbodiRSI \cite{N_EMBODIRSI} & Adaptive acquisition & Changes to the acquisition distribution and acquisition rules.\\[2pt]
RoboCoach \cite{N_ROBOCOACH} & Acquisition component & Component inheritance and subsequent learning gains.\\[2pt]
ENPIRE \cite{W028} & Policy-training programs & Training-code inheritance, capability gains, and fixed surrounding machinery.\\[2pt]
Eureka \cite{W052} & Reward design & Reward revisions and their role in subsequent policy training.\\[2pt]
STOP \cite{W109} & Optimization procedure & Optimizer inheritance and controls for subsequent gains.\\[2pt]
Darwin Godel Machine \cite{W111} & Agent programs & Inheritance of agent modifications and ablation controls.\\[2pt]
Dream-RSI \cite{W119} & Improvement procedure & Costs and reference conditions in improver comparisons.\\[2pt]
Escher-Loop \cite{W114} & Optimizer and search & Optimizer gains, task effects, and total search costs.\\[2pt]
PhysicalRSI 1.0 \cite{N_PHYSICALRSI} & Editable harness & Roles of inherited edits in learning and execution.\\[2pt]
\bottomrule
\end{tabular}
\end{table*}

\paragraph{Supervision and update objects.}
Similar headline descriptions can conceal different sources of autonomy. RoboCat and RECAP retain robot experience but organize supervision and policy updating differently. Q-Planning improves an off-policy critic while freezing its action generator, and retains human success labels in its physical experiments. Self-EFM supplies self-derived progress feedback but freezes a separate supervised-stage reward checkpoint during online policy improvement. Robot-generated experience, improved action selection, and self-derived evaluation thus identify different changed objects and remaining human roles. None is synonymous with an improving evaluator or learning algorithm. \cite{W005,W003,W010,W009}

\subsection{Benchmark design and evaluation scope}

Robot benchmarks define different evaluation objects. Meta-World separates multi-task competence from few-shot adaptation to held-out tasks. RLBench supplies a manipulation task collection with demonstrations and few-shot learning challenges; ManiSkill2 emphasizes manipulation skills across object and environment variation. Robomimic examines learning from offline human demonstrations, including data quality and algorithmic choices, while robosuite supplies reusable simulation and task infrastructure. These resources support capability and generalization studies, but a benchmark name alone does not identify an RSI evaluation. This paper distinguishes the actual protocol: episode, task chain, adaptation episode, sequential learning history, or inherited improver revision.\cite{F_METAWORLD,F_RLBENCH,F_MANISKILL2,F_ROBOMIMIC,F_ROBOSUITE}

Using a task environment to study self-improvement also requires specifying when learning occurs. LIBERO, RoboCasa, and RoboTwin provide task distributions and controlled conditions, ESI-Bench adds active embodied reasoning, and EmbodiedGovBench addresses operational constraints. A study can collect experience, update the system, and freeze the retained version for independent testing, or evaluate the entire stream while the system interacts and updates. These protocols characterize acquired capability and performance during online improvement, respectively. \cite{libero2023,robocasa2024,robotwin2025,W059,W060}

\paragraph{Task competence and retention.}
CALVIN tests long-horizon language-conditioned composition, making it useful for examining coherent action sequences. Longer completed chains, however, do not establish autonomous experience selection or improved subsequent learning. LIBERO directly addresses transfer in lifelong robot learning, including task order and pretraining. Current-task improvement and earlier-skill retention can diverge; aggregate success therefore needs per-task changes and exposure history. \cite{calvin2022,libero2023}

RoboCasa broadens household tasks and demonstration-generation infrastructure, exposing dependencies between objects and subgoals. A catalog of tasks is not evidence of autonomous competence across that catalog. RoboCasa365 separates multitask, foundation-model, and lifelong-learning evaluations. Its sequential fine-tuning experiment exhibits forgetting, while its tested mixture containing MimicGen demonstrations underperforms the human-only mixture. These are conditional findings about that data and training setup, not a general rejection of synthetic experience. Task diversity, data volume, and durable learning must be measured separately. \cite{robocasa2024,robocasa3652026}

\paragraph{Generation feedback and evaluation proxies.}
RoboTwin 2.0 uses simulation logs and visual diagnostics to revise generated task programs and produces trajectories under domain randomization. This supplies a feedback pathway and permits analysis of downstream data utility. Repairing one program, however, differs from improving the persistent procedure that generates programs. An RSI comparison must preserve and test that procedure across subsequent generation opportunities. \cite{robotwin2025}

SIMPLER reduces visual and control discrepancies between simulation and real-robot setups and compares paired evaluations. WorldGym combines an action-conditioned video model with a vision-language reward evaluator and reports agreement with physical outcomes and policy rankings in its evaluated setting. Such findings support calibrated screening within the examined regime. They do not establish calibration after optimizing a policy against the proxy, changing contact dynamics, or changing policy families. A plausible video can conceal a physical error also missed by its reward model; agreement between dependent models does not resolve that shared blind spot. \cite{simpler2024,worldgym2025}

\begin{figure*}[p]
\centering\includegraphics[width=\textwidth]{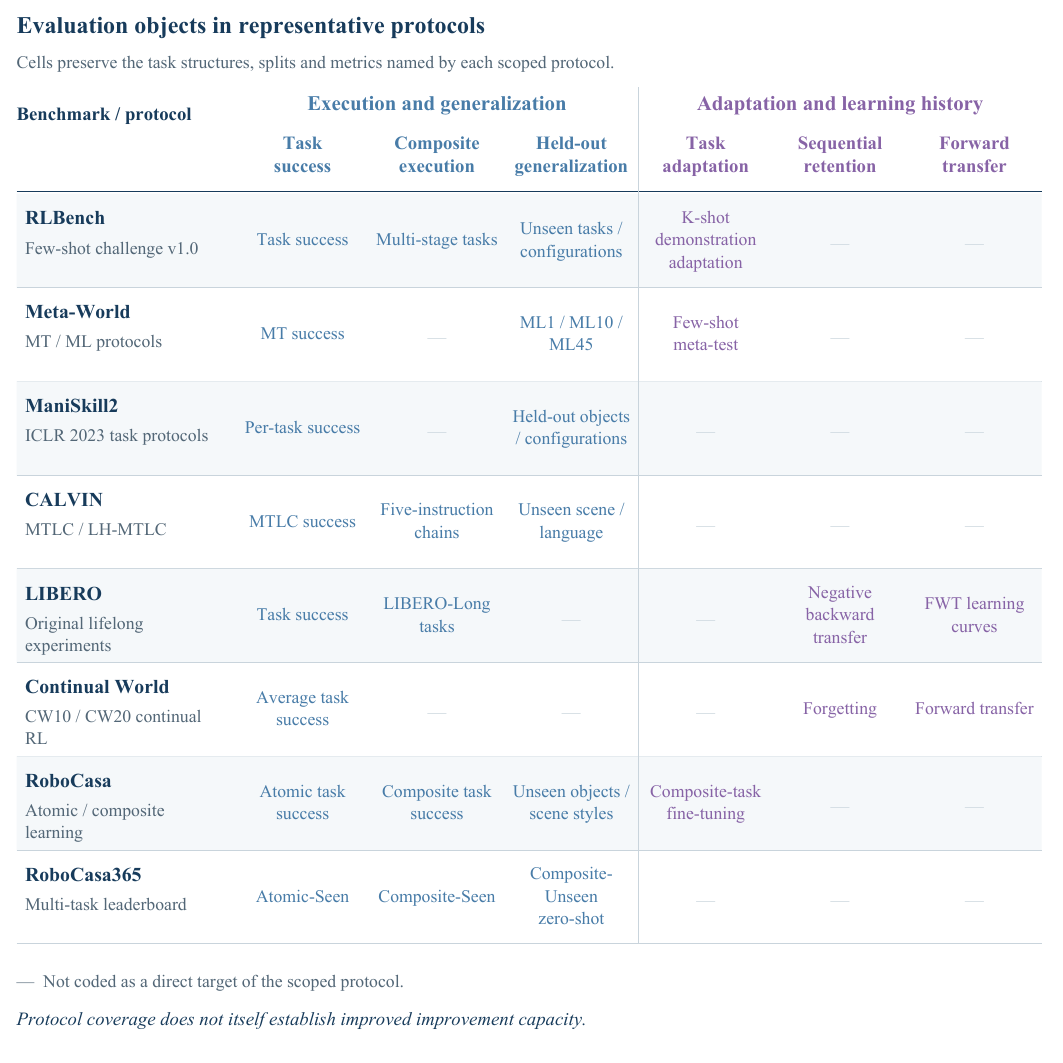}
\caption{Evaluation objects in representative robot-learning protocols. Cells name the task structure, split, or metric targeted by each scoped protocol. Dashes denote objectives not encoded as direct targets in that protocol, rather than capabilities absent from the environment. CALVIN instruction chains differ from a learning history; LIBERO and Continual World explicitly evaluate sequential retention and transfer. The RoboCasa365 row denotes its public multi-task leaderboard. Sources: RLBench, Meta-World, ManiSkill2, CALVIN, LIBERO, Continual World, RoboCasa, and RoboCasa365.\cite{F_RLBENCH,F_METAWORLD,F_MANISKILL2,calvin2022,libero2023,F_CONTINUALWORLD,robocasa2024,robocasa3652026}}
\label{fig:benchmarks}
\end{figure*}

Task structure and update protocol jointly determine the meaning of a benchmark result. Long action chains test continued execution, retesting across tasks measures retention, and comparing improvement procedures requires further learning opportunities. Figure~\ref{fig:benchmarks} relates these evaluation objects to representative protocols.

Digital sequential benchmarks provide useful protocol ideas. SEA-Eval tracks performance and process measures across tasks; SEAGym separates update-visible data from frozen validation and test conditions; SkillFlow studies whether accumulated skills are actually used and beneficial. Harness Updating Is Not Harness Benefit explicitly separates the ability to produce an update from the ability to exploit it. These distinctions motivate crossing improvers with learners rather than evaluating only their original pairings. \cite{W127,W128,W121,W123}

Negative evidence is essential to the synthesis. The autonomous-data-collection study shows why more self-collected experience cannot be assumed to outperform more demonstrations. What Stops Recursive Self-Improvement reports a bounded simulated robotics attempt in which extended iteration and human repairs did not establish the intended success. Self-Improvement Can Self-Regress and Scientific Amnesia document different forms of regression or failed accumulation in digital learning. These observations identify failure conditions; they neither refute every RSI mechanism nor justify classifying unsuccessful systems as L0. \cite{W061,N_WHATSTOPS,W126,W125}

A controlled improver comparison inherits the statistical and protocol problems already established in reinforcement learning. Deep Reinforcement Learning that Matters and Empirical Design in Reinforcement Learning motivate explicit reporting of implementations, selection procedures, seeds, and evaluation settings. Rliable supplies uncertainty-aware aggregation and performance-profile tools. Procgen separates training experience from held-out procedural environments, whereas D4RL fixes a dataset to study learning under a shared evidence supply. Together these works support two complementary RSI tests: compare how improvers use identical data, and compare what data they acquire under matched interaction budgets. Neither test alone isolates both acquisition and learning effects.\cite{F_RLIABLE,F_RLMATTERS,F_EMPIRICALRL,F_PROCGEN,F_D4RL}


\subsection{Controlled comparison of improvers}

Task benchmarks measure system performance, while before--after comparisons measure the effects of retained updates. Evaluating the improver requires starting the reference and revised procedures from the same learner and observing the subsequent gains each produces. We therefore consider two complementary conditions. A common-data test gives both improvers the same experience to compare its use. An interactive test allows each to collect experience under matched resource limits, comparing the joint effects of acquisition and updating. Shared foundation models, tool access, and independent test tasks help isolate the effect of the procedure change.

Freeze both improvers before the final comparison and keep their development and selection data separate from held-out tests. The shared learner start controls differences caused by co-adaptation in the original pairing. If further revisions are allowed during testing, predeclare the complete revision policy as part of the evaluated improver.

\begin{figure*}[t]
\centering\includegraphics[width=\textwidth]{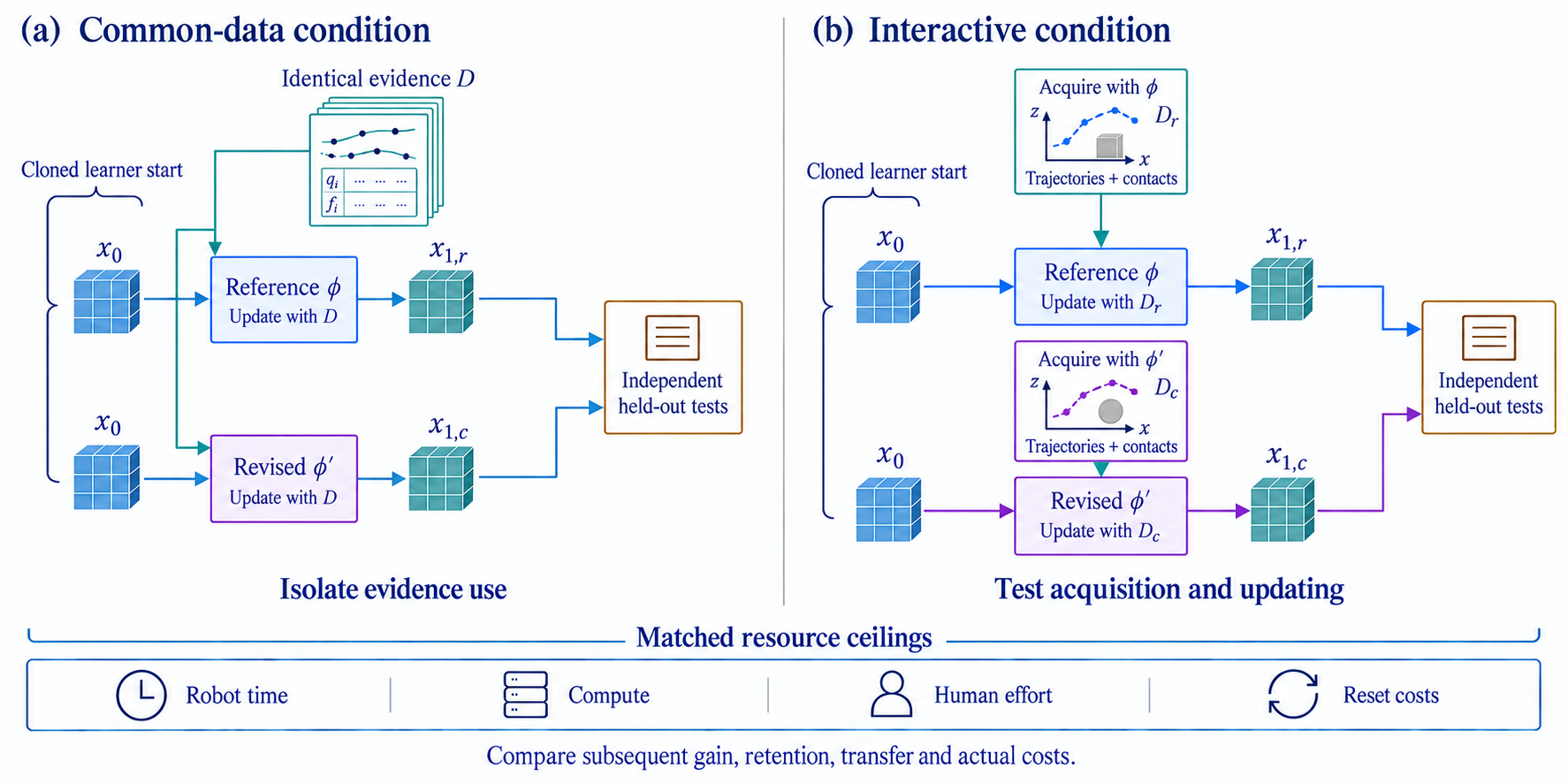}
\caption{Common-data and interactive conditions for improver comparison. In the common-data condition, cloned learner starts and identical permitted evidence enter the reference and revised improvers. In the interactive condition, each improver acquires its own evidence under matched starts and resource ceilings. Branch-specific successor learners are assessed by independent held-out tests that provide no update feedback. The two conditions assess evidence use and acquisition plus updating, respectively; report subsequent gain, retention, transfer, and actual costs.}
\label{fig:conditions}
\end{figure*}

\begin{figure*}[t]
\centering\includegraphics[width=0.90\textwidth]{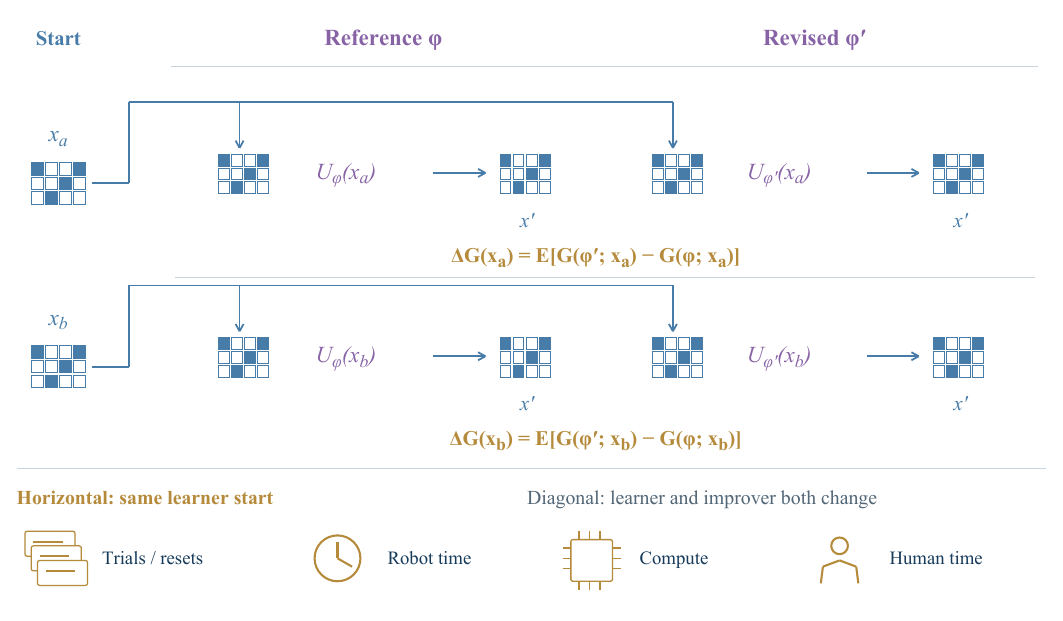}
\caption{Crossed evaluation of learner starts and improvement procedures. Each row clones a learner start and applies the reference and revised improvers; the horizontal contrast uses $G$ and $\Delta G$ as defined in Eq.~\eqref{eq:improver_gain}, with common budgets and test conditions omitted from the displayed notation. Diagonal comparisons change both learner state and improver. Estimate uncertainty over independent replicates and report absolute gains alongside relative contrasts. Match resource ceilings and record realized consumption, including construction, collection, computation, human assistance, energy, and resets. The diagram contains no measured results.}
\label{fig:swap}
\end{figure*}

This distinction preserves the value of acquisition itself. Common data controls experience differences when a revision changes optimization, while interactive conditions allow a revised exploration strategy or curriculum to influence the data collected. Figure~\ref{fig:conditions} contrasts these conditions, and Figure~\ref{fig:swap} shows the pairing of learner starts with improver versions.

The target quantity is subsequent gain, measured relative to the shared starting learner. An improver comparison should report the difference between those gains across independent runs or task groups, with uncertainty and actual resource use. Fixed-budget final performance, area under a learning curve, and cost to reach a predefined threshold answer different questions and should not be conflated. Improvement rate is not established by dividing a single final score by an arbitrary number of iterations.

\begin{align}
G(\varphi;x_0,B,\mathcal T)
 &= J_{\mathcal T}\!\left(U_{\varphi}^{B}(x_0)\right)-J_{\mathcal T}(x_0),\\
\Delta G &= \mathbb E\!\left[G(\varphi';x_0,B,\mathcal T)
                    -G(\varphi;x_0,B,\mathcal T)\right].
\label{eq:improver_gain}
\end{align}
Here, $U_{\varphi}^{B}$ denotes the complete improvement run within resource limits $B$, and $J_{\mathcal T}$ is an independently assessed outcome on held-out tasks $\mathcal T$. The expectation is over the declared starts, task groups, and stochastic runs. A positive estimate of $\Delta G$ supports an improver-gain claim only with uncertainty, independence, and budget controls; this expression defines the evaluation target. A positive difference can occur even when both learners regress. Report absolute gains separately and require positive gain of the revised process, as well as its relative advantage, before claiming beneficial recursive improvement.

Where interfaces permit, cross the improvers with multiple initial learners and cross the update procedures with evidence acquired by either system. A revised improver may only work with its co-adapted learner, or may benefit because it acquired a more favorable dataset. These are useful findings, but they are narrower than a reusable improvement advantage. If evidence cannot be exchanged because sensors, actions, or morphology differ, compare compatible subsets and state the restriction instead of treating a failed swap as evidence against the mechanism.

\paragraph{Lineage and reference conditions.}
A positive $\Delta G$ establishes a conditional advantage for the tested starts, tasks, and resources. Attribution to recursion additionally requires versioned records showing that the system generated, inherited, and used the revised improver. A manually engineered replacement can improve learning without establishing that provenance. First hold the acquired improver fixed during the measured opportunity; permitting further self-modification defines a separate total-effect condition. Add a frozen learner to detect environmental or measurement drift, and an evolving learner with a frozen improvement procedure to represent ordinary learning. Repeated checkpoints test whether the advantage survives reuse.

Forcing identical trajectories suppresses any benefit of improved exploration; unrestricted acquisition confounds an optimizer-only comparison. The difference between common-data and interactive effects is not automatically an identified mediation effect, because acquisition and updating can interact and visit different states. For physical runs, randomize assignment within appropriate task, hardware, site, and time blocks. Count aborted trials and failed resets using rules fixed in advance. Returning a robot to the same pose may not restore object condition, actuator temperature, or other relevant physical state.

\paragraph{Resources, information flow, and uncertainty.}
Treat the budget as a vector with explicit units, for example
\begin{equation}
\begin{aligned}
\mathbf b=(&n_{\mathrm{trans}},t_{\mathrm{robot}},c_{\mathrm{train}},\\
          &c_{\mathrm{verify}},t_{\mathrm{human}},e_{\mathrm{energy}},n_{\mathrm{reset}}).
\end{aligned}
\end{equation}
Transition counts, robot time, training and verification compute, human time, energy, and resets are not interchangeable. Match componentwise ceilings and report actual consumption. Candidate discovery, rejected updates, recovery, and maintenance belong in the end-to-end record. Report improver construction separately from recurring use, with explicit amortization assumptions. When exact matching is impossible, compare a prespecified budget grid and performance--resource frontier. Any scalar conversion needs stated rates and sensitivity analysis.

Separate development feedback from concealed reporting audits. Checkpoint scores that influence subsequent updates become development information for those later rounds, even if the underlying episodes remain hidden. Preserve a terminal holdout until the improvement procedure and analysis decisions are frozen. Version task generators, models, code, intervention records, and access permissions so that the evidence available to each updater can be reconstructed.

Uncertainty must follow the assignment structure. Episodes sharing a task, robot, site, or seed are not automatically independent replicates; analyze paired contrasts within blocks and use cluster-aware uncertainty where appropriate. Report old-skill changes against a fixed anchor alongside new-task gains, because acceptable average retention can conceal loss of an essential skill. Safety conclusions require an event definition, exposure, and interruption policy: zero observed incidents is not zero risk, and shielded trials evaluate the shield--learner combination. Intervention frequency is needed to interpret apparent safety and autonomy. Table~\ref{tab:improver-controls} brings these requirements together, linking each comparison condition to the information needed to interpret an improver-gain claim.

\begingroup\small
\setlength{\tabcolsep}{3pt}
\begin{table}[tb]
\centering\fontsize{8.5}{10.5}\selectfont
\caption{Controls for an improver-gain claim}
\label{tab:improver-controls}
\begin{tabular}{>{\raggedright\arraybackslash}p{0.30\columnwidth}>{\raggedright\arraybackslash}p{0.62\columnwidth}}
\toprule
\textbf{Control} & \textbf{Purpose and required report}\\\midrule

Shared learner starts & Match model version, initial weights, retained skills, prior data, and accessible tools.\\[3pt]
Common-data condition & Give both improvers identical admissible evidence; isolate how the evidence is used.\\[3pt]
Interactive condition & Allow separate collection under matched physical and compute limits; count unsuccessful acquisition.\\[3pt]
Independent assessment & Separate development feedback from held-out outcomes; freeze the assessor or calibrate versions externally.\\[3pt]
Retention and transfer & Test prior tasks, new tasks, and compatible alternative learners; report negative transfer.\\[3pt]
Repeated runs and costs & Report uncertainty, full trajectories, total search cost, human assistance, resets, and maintenance.\\[3pt]
\bottomrule
\end{tabular}
\end{table}
\endgroup


\subsection{Quantitative results for policies, memory, and system evolution}

Quantitative results show how these mechanisms operate in different tasks. Manipulation policies are typically evaluated by success rate, while memory and harness methods additionally address partially observed state, long task sequences, and reuse across tasks. We compare LIBERO, EmbodiedBench, RoboMemArena, and RoboDojo separately and interpret gains using each study's initial system and resource conditions. Benchmark scores describe the evaluated configuration; within-study before-and-after contrasts more directly show the change associated with an improvement.

EmbodiedBench supports several distinct improvement routes. RoboFoundry revises the executable agent during the evaluated stream, whereas EmbodiSkill develops reusable skills before held-out testing. MemCompiler trains state-dependent memory use, and ERA updates a compact policy through embodied prior learning and online reinforcement learning. Their reported scores locate these mechanisms within task benchmarks; the update boundary and training resources determine which comparisons isolate an improvement effect.\cite{N_ROBOFOUNDRY,W041,N_MEMCOMPILER,N_ERA}

Memory-dependent manipulation involves both retained state and reliable execution. Table~\ref{tab:robomemarena} compares memory-augmented VLAs, WAM execution controls, HarnessWAM, and RoboFoundry on the full 26-task benchmark. HarnessWAM tests structured state maintenance and recovery during execution, linking task performance to how memory is used.

HyMeS evaluates a learned executable memory heuristic that is frozen at test time. On its selected 12-task evaluation (160 episodes), it reports overall TSR and CSR of 60.1\% and 66.2\%, compared with 45.6\% and 61.7\% for the re-evaluated PrediMem baseline. This within-study comparison uses corrected success checks and a selected task set, distinct from the full-benchmark results.\cite{N_HYMES}

\begin{table*}[tp]
\caption{Reported task and cumulative success on the full 26-task RoboMemArena benchmark (\%).}
\label{tab:robomemarena}
\centering
\fontsize{9}{11}\selectfont
\setlength{\tabcolsep}{4pt}
\renewcommand{\arraystretch}{1.18}
\begin{tabular*}{\textwidth}{@{\extracolsep{\fill}}lrrrrrrrrrr@{}}
\toprule
 & \multicolumn{2}{c}{\textbf{Transfer}} & \multicolumn{2}{c}{\textbf{Occlusion}} & \multicolumn{2}{c}{\textbf{Counting}} & \multicolumn{2}{c}{\textbf{Sequence}} & \multicolumn{2}{c}{\textbf{Overall}}\\
\cmidrule(lr){2-3}\cmidrule(lr){4-5}\cmidrule(lr){6-7}\cmidrule(lr){8-9}\cmidrule(lr){10-11}
\textbf{Method} & TSR & CSR & TSR & CSR & TSR & CSR & TSR & CSR & TSR & CSR\\
\midrule
$\pi_{0.5}$ & 20.0 & 42.8 & 12.7 & 17.2 & 14.3 & 50.9 & 60.0 & 71.6 & 21.5 & 38.7\\
HiF-VLA & 17.5 & 38.9 & 12.7 & 27.1 & 8.6 & 45.9 & 42.5 & 70.2 & 16.9 & 39.8\\
MemoryVLA & 15.0 & 37.2 & 7.3 & 13.1 & 14.3 & 55.1 & 37.5 & 65.2 & 15.0 & 35.3\\
MemER & 20.0 & 36.1 & 16.4 & 33.2 & 27.1 & 65.1 & 65.0 & 79.1 & 27.3 & 49.1\\
PrediMem \cite{N_ROBOMEMARENA} & 22.5 & 45.2 & 27.3 & 38.4 & 45.7 & 69.3 & 72.5 & 89.5 & 38.5 & 55.2\\
\midrule
WAM (whole task) & 11.3 & 12.5 & 41.3 & 42.8 & 51.4 & 75.8 & 73.8 & 77.1 & 44.4 & 52.3\\
WAM (static plan) & 15.0 & 16.7 & 38.2 & 49.8 & 67.2 & 90.1 & 73.8 & 91.9 & 47.9 & 62.0\\
HarnessWAM \cite{N_HARNESSWAM} & 21.3 & 31.5 & 55.0 & 63.9 & 73.6 & 88.2 & 86.3 & 93.0 & 59.6 & 69.9\\
\midrule
RoboFoundry \cite{N_ROBOFOUNDRY} & 70.5 & 78.1 & 38.3 & 57.6 & 55.2 & 82.5 & 75.2 & 92.1 & 53.5 & 72.8\\
\bottomrule
\end{tabular*}
\par\nobreak\vspace{7pt}
\noindent\begin{minipage}{\textwidth}
\fontsize{8.5}{11}\selectfont\raggedright
Sources: RoboMemArena, Table~2 \cite{N_ROBOMEMARENA}; HarnessWAM, Table~1 \cite{N_HARNESSWAM}; RoboFoundry, Table~2 \cite{N_ROBOFOUNDRY}. TSR measures complete task success; CSR measures completed verification stages. Overall values preserve each source's benchmark aggregation rather than averaging the four displayed categories equally. The WAM controls and HarnessWAM share the LingBot-VA checkpoint and use 20 rollouts per task. RoboFoundry uses Qwen3.7-Plus and counts evolution episodes in its task stream. Cross-paper resource budgets and update protocols differ.
\end{minipage}
\end{table*}

\begin{table*}[tp]
\caption{Reported RoboDojo simulation results from the 28 September 2026 leaderboard snapshot.}
\label{tab:robodojo-performance}
\centering
\fontsize{9}{11}\selectfont
\setlength{\tabcolsep}{4pt}
\renewcommand{\arraystretch}{1.18}
\begin{tabularx}{\textwidth}{@{}Ylrrrrrrrr@{}}
\toprule
& & \multicolumn{6}{c}{\textbf{Category score}} & \multicolumn{2}{c}{\textbf{Average}}\\
\cmidrule(lr){3-8}\cmidrule(lr){9-10}
\textbf{Method} & \textbf{Type} & \textbf{Gen-S} & \textbf{Gen-R} & \textbf{Prec.} & \textbf{Long} & \textbf{Mem.} & \textbf{Open} & \textbf{Score} & \textbf{SR (\%)}\\
\midrule
PhysicalRSI 1.0 & Agent+VLA & 31.57 & 11.04 & 38.05 & 46.37 & 46.74 & 28.88 & 36.27 & 31.38\\
Simate-beta & VLA & 40.54 & 29.63 & 34.35 & 57.84 & 33.33 & 9.12 & 33.95 & 27.96\\
VPP2-Preview & WAM & 43.92 & 18.75 & 33.75 & 35.31 & 50.78 & 5.83 & 31.40 & 25.62\\
Liber-0 Preview & VLM+WAM & 39.31 & 10.67 & 38.28 & 45.98 & 37.77 & 6.68 & 30.74 & 25.52\\
GPT-6 Astra & Agent & 35.32 & 31.40 & 12.65 & 21.45 & 43.04 & 34.36 & 28.97 & 22.48\\
DM0.5 & VLA & 23.49 & 8.06 & 24.82 & 33.70 & 47.74 & 2.43 & 24.90 & 19.34\\
\bottomrule
\end{tabularx}
\par\nobreak\vspace{7pt}
\noindent\begin{minipage}{\textwidth}
\fontsize{8.5}{11}\selectfont\raggedright
Source: the XPolicyLab mirror \cite{N_XPOLICY_BOARD}. Score is progress on a 0--100 scale; SR is terminal success. Gen-S/R denote standard/randomized generalization. Published averages and submission settings are retained; this is a dated snapshot, not a matched-budget improver experiment.
\end{minipage}
\end{table*}

\begin{table*}[tp]
\caption{Improvement mechanisms and evaluation scope of RoboDojo studies.}
\label{tab:robodojo-evolution}
\centering
\fontsize{9}{11}\selectfont
\setlength{\tabcolsep}{4pt}
\renewcommand{\arraystretch}{1.18}
\begin{tabularx}{\textwidth}{@{}P{0.16\textwidth}P{0.12\textwidth}P{0.19\textwidth}YY@{}}
\toprule
\textbf{Method} & \textbf{Coverage} & \textbf{Retained artifact} & \textbf{Revision and testing} & \textbf{Evidence focus}\\
\midrule
PhysicalRSI 1.0 \cite{N_PHYSICALRSI} & Simulation & Harness and skills & Candidate revisions are tested, selected and inherited. & System evolution and task performance.\\[3pt]
HexaAnything \cite{N_HEXA} & 3 tasks & Tool code & Two revision rounds; fixed evaluator. Cloth trials include human-selected action points. & Direct tool-revision trajectory.\\[3pt]
URAI \cite{N_URAI} & 5 tasks & Tool code & Human-guided development; tools frozen for four execution models. & Tool use and a separate revision case.\\[3pt]
ENCORE \cite{N_ENCORE} & 22 tasks & Policy programs & Demonstration-guided refinement; programs frozen before sealed tests. & Program discovery and generalization.\\[3pt]
RoboHarn-Evo \cite{REC_RoboHarnEvo2026} & 2 tasks & Physical knowledge & Knowledge evolves on RMBench and remains read-only on RoboDojo. & Transfer of evolved knowledge.\\[3pt]
ME-Brain-1.0 \cite{N_MEBRAIN} & 54 configurations & Memory and skills & RoboDojo evaluates the action model; multi-round evolution is not isolated. & Task performance of an evolvable system.\\
\bottomrule
\end{tabularx}
\par\nobreak\vspace{7pt}
\noindent\begin{minipage}{\textwidth}
\fontsize{8.5}{11}\selectfont\raggedright
Coverage refers to each paper's evaluated subset; configurations are not distinct task definitions. Performance, inherited revisions and greater subsequent learning require different controls. Sources: HexaAnything \S6.3; URAI \S4; ENCORE \S4.5; RoboHarn-Evo \S5.4; ME-Brain \S7.2.2.
\end{minipage}
\end{table*}

LIBERO results show that policy gains can be obtained through different interfaces. RedFlow converts fixed failure experience into corrections, increasing average success from 56.2\% to 68.2\% in its setting. Q-Planning keeps the FastWAM action generator frozen and improves action selection, raising the four-suite mean from approximately 94.4\% to 99.1\% in its online setting. Q-VGM separates offline and online updates, while WoVR supports policy training with model-generated experience. These contrasts illustrate the roles of failure reuse, candidate evaluation, and imagined training. Different initial models and collection budgets explain why absolute scores or gain magnitudes cannot replace mechanism comparisons.\cite{W025,W026,W010,W017}\par
For system evolution, evaluation timing also affects interpretation. RoboFoundry's streaming results include performance during adaptation, whereas a held-out test after freezing updates measures retained capability. The former describes improvement in operation; the latter describes the system obtained after improvement. \cite{N_ROBOFOUNDRY}

\begin{table*}[tp]
\caption{Within-study performance changes after improvement.}
\label{tab:reported-gains}
\centering
\fontsize{9}{11}\selectfont
\setlength{\tabcolsep}{4pt}
\renewcommand{\arraystretch}{1.18}
\begin{tabularx}{\textwidth}{@{}l P{0.24\textwidth}Y rrr@{}}
\toprule
\textbf{Method} & \textbf{Evaluated scope} & \textbf{Changed component} & \textbf{Reference} & \textbf{Updated} & \textbf{$\Delta$ (pp)}\\
\midrule
EmbodiRSI \cite{N_EMBODIRSI} & S1+S3 simulation & Data acquisition & 59.3 & 79.9 & +20.6\\
FIND \cite{N_FIND} & 8 Franka tasks & Residual policy & 55.0 & 71.9 & +16.9\\
RoboCoach \cite{N_ROBOCOACH} & 3 Franka tasks & Skill policies & 13.3 & 75.0 & +61.7\\
Motus2 \cite{W002} & 2 real-robot tasks & Action parameters & 65.0 & 72.5 & +7.5\\
LWD \cite{W004} & Freezer restocking & Fleet policy & 92.0 & 97.0 & +5.0\\
RPG \cite{N_RPG} & 4 simulation tasks & Skill library & 65.0 & 92.5 & +27.5\\
SHAPER \cite{W029} & ESI-Bench, 231 questions & Execution harness & 41.1 & 49.8 & +8.7\\
\bottomrule
\end{tabularx}
\par\nobreak\vspace{7pt}
\noindent\begin{minipage}{\textwidth}
\fontsize{8.5}{11}\selectfont\raggedright
Values are SR (\%), except SHAPER's micro accuracy (\%). $\Delta$ is the within-row difference in percentage points. Reference conditions, update resources and test designs are specified in Table~\ref{tab:resource-protocols}; different task sets are not ranked.
\end{minipage}
\end{table*}

\begin{table*}[tp]
\caption{Data, update conditions and evaluation resources for the reported contrasts.}
\label{tab:resource-protocols}
\centering
\fontsize{9}{11}\selectfont
\setlength{\tabcolsep}{4pt}
\renewcommand{\arraystretch}{1.18}
\begin{tabularx}{\textwidth}{@{}l P{0.28\textwidth}YY@{}}
\toprule
\textbf{Method} & \textbf{Reference and update setting} & \textbf{Experience or resource budget} & \textbf{Evaluation}\\
\midrule
Q-VGM & Few-shot SFT; offline then online & 150 offline rollouts/suite & 500 tests/suite; 3 training seeds\\[2pt]
RedFlow & Base policy; one offline update & 1,536 rollouts/suite & 500 tests/suite; horizons 240/240/240/480\\[2pt]
Q-Planning & Frozen FastWAM; learned Q & 100 rollouts/task/round; 10 rounds & 20 tests/task; horizon 540 for every row\\[2pt]
WoVR & Full-trajectory SFT; imagined RL & 2,500 simulator trajectories/suite & Four LIBERO suites\\
\midrule
EmbodiRSI & Static versus adaptive acquisition & 400 retained trajectories/subtask at each endpoint & S1+S3 scene-balanced SR; no HIL\\[2pt]
FIND & Base versus online residual policy & 50 initial demonstrations/task & 5 runs; 20 independent tests/task/checkpoint\\[2pt]
RoboCoach & Initial versus coached skills & 150 additional human demonstrations in 3 rounds & 20 tests/task/round\\[2pt]
Motus2 & Same checkpoint; no planning & MBRL changes action parameters only & 20 tests/task\\[2pt]
LWD & Offline versus online fleet policy & 4 h; about 60 robot-hours across 16 robots/8 tasks & Freezer episode SR\\[2pt]
RPG & Initial versus repaired library & Model, prompt, perception and test budget fixed & 10 paired starts/task; 40 episodes\\[2pt]
SHAPER & Same evolved skills; revised harness & 10 development and 10 validation questions & 231 held-out questions; 30-step cap\\
\bottomrule
\end{tabularx}
\par\nobreak\vspace{7pt}
\noindent\begin{minipage}{\textwidth}
\fontsize{8.5}{11}\selectfont\raggedright
LIBERO sources are Q-VGM, RedFlow, Q-Planning, and WoVR.\cite{W025,W026,W010,W017} References for the other methods are given in Table~\ref{tab:reported-gains}. Source tables: Q-VGM II; RedFlow 1; Q-Planning 3; WoVR 2; EmbodiRSI 1; FIND Fig.~3(a) and I--II; RoboCoach 12; Motus2 3; LWD I; RPG IV; SHAPER 3. Equal retained data need not imply equal acquisition cost. LWD cost is fleet-wide. RoboCoach's uniform-acquisition control reaches 30.0; Motus2's planning-only control reaches 67.5. Full cost accounting remains source-dependent.
\end{minipage}
\end{table*}

Real-robot and system-level contrasts further connect gains to resources. FIND increases success from 55.0\% to 71.9\% across eight Franka tasks. LWD improves freezer restocking from 92.0\% to 97.0\% using online fleet experience. EmbodiRSI and RoboCoach change experience allocation, whereas RPG and SHAPER change reusable execution artifacts (Table~\ref{tab:reported-gains}). These results support capability improvement within different scopes. Retained trajectories, human demonstrations, fleet time, and test-time action limits describe different acquisition and execution costs; Table~\ref{tab:resource-protocols} gives the corresponding conditions.

RoboDojo supports both benchmark-wide submissions and studies of selected tasks. Table~\ref{tab:robodojo-performance} preserves a dated leaderboard snapshot: its category scores and terminal success rates summarize task performance. Table~\ref{tab:robodojo-evolution} shows why such results need a separate mechanism-level reading. HexaAnything and URAI examine tool revision with different forms of human participation; ENCORE evaluates discovered programs after development; RoboHarn-Evo transfers knowledge evolved on another benchmark. The action-model results of ME-Brain-1.0 assess a different part of its system. These protocols provide evidence about revision, transfer, or task competence, while a reusable improver advantage requires the controlled comparisons described above.\cite{N_HEXA,N_URAI,N_ENCORE,REC_RoboHarnEvo2026,N_MEBRAIN}

\subsection{Synthesis of the current evidence}

The approaches establish multiple routes from experience to persistent capability: practice updates policies, prediction expands usable experience, skill libraries retain reusable behaviors, and harnesses organize their invocation. Revisions to rewards and training programs extend change to the process that produces capability. The common connection is that an interaction can influence later tasks and even the organization of later learning. Benchmarks and within-study contrasts reveal different parts of this relationship; improver comparisons from shared learner starts target the subsequent value of the process revision itself.

Digital-agent research contributes designs for this latter comparison. Dream-RSI and Escher-Loop evaluate programs that produce later solutions, while DGM analyzes system modification through related ablations. Together with robotic policy, memory, and harness results, these studies provide methodological references. Further development of physical RSI connects reusable process revisions to physical interaction budgets, recovery, and retention, testing whether their advantages accumulate during continued deployment. \cite{W119,W114,W111}

\section{Open Challenges and Research Directions}
\label{sec:directions}

When experience acquisition, feedback, and learning procedures can all be revised, sustained improvement also depends on their remaining compatible as they change together. The preceding comparisons direct attention to this continuity: feedback must remain grounded in physical outcomes, acquired capabilities and improvement procedures must survive further updates and transfer, and repeated experimentation requires reliable recovery at an affordable resource cost. Establishing progress also requires records of process versions, interventions, and evaluation conditions so that gains can be reproduced and attributed. The following research directions develop these connected requirements.

\subsection{Grounded feedback that survives system change}

The first priority is to maintain a reliable connection between internal feedback and physical outcomes as the system changes. Reward models need failure coverage, temporal context, calibration, and audits outside their training loop. Proposals such as intrinsic robot rewarding can contribute ideas, but proposed protocols must be distinguished from completed online experiments. A useful research target is not merely a higher reward-model score, but a feedback mechanism whose errors remain bounded under the distribution induced by continued improvement. \cite{W053,W055,W058}

\subsection{Lasting gains and reusable improvers}

A useful improvement should survive later updates and transfer beyond the failure that triggered it. This requires studying skill retirement, conflicting memories, representation drift, and compatibility between a revised improver and different learners. SkillSmith and Evo-Harness suggest ways to manage external artifacts, while morphology-design systems expose transfer across changing bodies. Physical studies should test whether a learned acquisition, verification, or update strategy benefits a new task or embodiment after its own development feedback is frozen. \cite{W116,W117,W082}

\subsection{Recovery and resource-aware experimentation}

Recovery should become part of the improvement objective rather than invisible laboratory labor. Systems need to reason about recoverable failure states, when to request assistance, and when an experiment should stop. Automated laboratories provide an adjacent physical setting in which experiments have delayed outcomes and material costs. They can inform provenance and scheduling, but successful automated synthesis does not itself establish an improving general-purpose robot or an improving research algorithm. \cite{W037,W062,W079}

Resource-aware search must include the cost of discovering the improver, not only the cost after discovery. Cheap simulated search may shift work to model construction, and fewer robot trials may require more demonstrations or human maintenance. Future benchmarks should expose these trade-offs as resource profiles rather than collapse them into one score. They should also include unsuccessful campaigns so that efficiency is not estimated only among survivors.

\subsection{Reproducible process evidence and review limitations}

Research should publish versioned learner and improver artifacts, acceptance decisions, intervention logs, and independent evaluation partitions. The verification gap in autonomous research and guidance-sensitive benchmarks such as ASI-Bench emphasize that finishing an experimental workflow, producing a plausible paper, and establishing a reproducible finding are different outcomes. Physical RSI needs the same distinction, augmented with hardware state and recovery records. \cite{W098,W129}

A concrete research target is an interchangeable improvement interface specifying inputs, editable components, admissible evidence, and resource limits. This would enable cross-learner tests without undocumented engineering changes. Cross-embodiment transfer should distinguish reuse of semantic knowledge, acquisition strategy, optimization code, and control representation; extensive human retuning changes the claim. Development evaluators may evolve, but the reporting outcome must retain a common interpretation. Changing success thresholds, task exclusions, or failure definitions can otherwise mimic progress without improving competence or improvement capacity.

The comparisons in this paper follow the retained artifacts, reset boundaries, and resource conditions reported for each evaluation protocol. Differences in task distributions, human assistance, and cost accounting limit direct comparisons across studies. The boundary between learner and improver follows a component's actual role in subsequent updates. A shared annotation protocol and inter-rater agreement tests can further improve the reproducibility of comparisons using the L0--L3 rubric.

\FloatBarrier
\section{Conclusion}
\label{sec:conclusion}

Embodied self-improvement rests on turning interaction experience into capability that remains useful later. Policies, world models, memory, skills, and harnesses provide different routes for retaining and reusing that experience. Recursive self-improvement further changes how data are acquired, feedback is constructed, and updates are produced, allowing inherited process revisions to govern the next learning round. This relationship connects otherwise separate technical approaches while preserving their differences in supervision, update objects, and physical conditions.

This paper relates capability gains, process inheritance, and improver gains through the L0--L3 evidence rubric and compares their mechanisms and experimental results. Long-term progress in embodied RSI depends jointly on valid feedback, continued interaction, retention, and affordable improvement costs. Common-data and interactive comparisons of revised improvers can further distinguish a system that has acquired better task capability from one that has acquired a more effective process for developing capability.


\FloatBarrier
\bibliographystyle{IEEEtran}
\bibliography{references}
\EOD
\end{document}